\documentclass{article}
\usepackage{ijcai22}

\usepackage{times}
\usepackage{soul}
\usepackage{url}
\usepackage{thm-restate}
\usepackage{thmtools} 

\usepackage[hidelinks]{hyperref}
\usepackage[utf8]{inputenc}
\usepackage[small]{caption}
\usepackage{graphicx}
\usepackage{amsmath}
\usepackage{amsthm}
\usepackage{booktabs}
\usepackage{algorithm}
\usepackage{mleftright}
\usepackage{bm}
\usepackage{xcolor}
\usepackage{amsmath}
\usepackage{amsthm}
\usepackage{amsfonts}
\usepackage{bbm}
\usepackage{mathtools}
\usepackage{dsfont}

\usepackage{pgfplots}
\pgfplotsset{compat=1.15}
\usepackage{mathrsfs}
\usetikzlibrary{arrows}
\usepackage[noend]{algpseudocode}
\usepackage{algpseudocode}
\usepackage{amssymb}
\usepackage{mathtools}
\usepackage{amsmath}
\usepackage{tikz}
\usepackage{pgfplots}
\usetikzlibrary{shapes.geometric}
\pgfplotsset{compat=newest}
\usepgfplotslibrary{groupplots}
\usepgfplotslibrary{dateplot}
\usepackage{todonotes}
\usepackage{dsfont}
\usepackage{enumitem}
\usepackage{tikz}
\usepackage{thm-restate}
\usepackage{booktabs}
\usepackage{multirow}
\usepackage{newfloat}
\usepackage{wrapfig}
\usepackage{dsfont}
\usepackage{comment}
\usepackage{multirow}
\usepackage{mleftright}
\usepackage{bm}
\usepackage{comment}
\usepackage{bbm}
\usepackage{subfig}
\usepackage{placeins}
\usepackage{makecell}

\usepackage{bm}
\usepackage{cancel}

\newtheorem{example}{Example}
\newtheorem{theorem}{Theorem}

\newtheorem{definition}{Definition}

\DeclareMathOperator*{\argmax}{argmax}

\newcommand{\R}{\mathcal{R}} 
\newcommand{\Tmax}{\tau_{\max}}

\newcommand{\step}{\phi}

\allowdisplaybreaks

\title{Multi-Armed Bandit Problem with Temporally-Partitioned Rewards: \\ When Partial Feedback Counts}

\author{
	Giulia Romano\and
	Andrea Agostini\and
	Francesco Trov\`o\and
	Nicola Gatti\And
	Marcello Restelli
	\affiliations
	Politecnico di Milano, Piazza Leonardo da Vinci 32, I-20133, Milan, Italy 
	\emails
	\{giulia.romano, francesco1.trovo, nicola.gatti, marcello.restelli\}@polimi.it, andrea1.agostini@mail.polimi.it
}

\begin{document}

\maketitle

\begin{abstract}
\textcolor{black}{
There is a rising interest in industrial online applications where data becomes available sequentially.
	%
%
Inspired by the recommendation of playlists to users where their preferences can be collected during the listening of the entire playlist, we study a novel bandit setting, namely \emph{Multi-Armed Bandit with Temporally-Partitioned Rewards} (TP-MAB), in which the stochastic reward associated with the pull of an arm is partitioned over a finite number of consecutive rounds following the pull.
This setting, unexplored so far to the best of our knowledge, is a natural extension of delayed-feedback bandits to the case in which rewards may be dilated over a finite-time span after the pull instead of being fully disclosed in a single, potentially delayed round.	
%
%
We provide two algorithms to address TP-MAB problems, namely, \texttt{TP-UCB-FR} and \texttt{TP-UCB-EW}, which exploit the partial information disclosed by the reward collected over time.
We show that our algorithms provide better asymptotical regret upper bounds than delayed-feedback bandit algorithms when a property characterizing a broad set of reward structures of practical interest, namely \emph{$\alpha$-smoothness}, holds.
We also empirically evaluate their performance across a wide range of settings, both synthetically generated and from a real-world media recommendation problem.
}
\end{abstract}

\section{Introduction} \label{S:intro}
\textcolor{black}{
Sequential decision-making occurs in many real-world scenarios such as clinical trials, recommender systems, web advertising, and e-commerce.
Inspired by these applications, many different flavours of the multi-armed bandit (MAB) setting have been investigated.
A crucial role is played by the time the reward is observed.
In many cases, the reward is subject to a \emph{delay}, and such a delay, if not sufficiently short, can prevent the design of algorithms that are effective in practice. 
Online learning with delayed feedback has received considerable attention in recent years, and several results are available in the literature, \emph{e.g.}, see the seminal work by~\citeauthor{joulani2013online}~[\citeyear{joulani2013online}].
%
%
%
A major distinction in MABs with delayed feedback concerns the nature of the rewards, which may be stochastic \cite{mandel2015queue,cella2020stochastic} or adversarial~\cite{bistritz2019exp3,thune2019nonstochastic,van2021nonstochastic}. 
}

\textcolor{black}{
Our work focuses on a special class of bandit problems with stochastic and delayed rewards, in which we can get partial feedback over time.
%
%
%
More precisely, we study a novel setting, namely MAB with Temporally-Partitioned Rewards (TP-MAB), in which the reward associated with an action, a.k.a.~\emph{arm}, chosen at a given round is collected during a finite number of rounds following the choice, according to an unknown probability distribution. 
In classical delayed-feedback bandits (see, \emph{e.g.},~\citeauthor{joulani2013online}~[\citeyear{joulani2013online}]), the reward is concentrated in a single round that is (stochastically) delayed w.r.t.~the round in which the learner pulled the corresponding arm.
TP-MABs naturally extend this setting by allowing the reward to be partitioned into multiple elements that are collected with different delays.
%
%
We call arm's \emph{per-round reward} the partial reward observed by the learner in a single round, which is assumed to be the realization of a random variable with an unknown probability distribution.
We call arm's \emph{cumulative reward} the random variable given by the sum of all the per-round rewards obtained by pulling an arm.
While the per-round reward can be observed round by round, the cumulative reward is revealed only at the end.
Notice that, in a single round, the learner observes a per-round reward for each previously pulled arm whose cumulative reward is not terminated yet.
Our goal is to find a policy to maximize the cumulative reward, exploiting the per-round rewards as intermediate signals on the arm performance.
}

\paragraph{Motivating applications.}
\textcolor{black}{
A motivating example for TP-MABs is recommending media content and, in particular, song playlists to a class of users (\emph{i.e.}, users sharing similar characteristics).
%
In this setting, each arm corresponds to a playlist.
The reward is measured in listening time (proportional to the user's appreciation).
The goal is to find the playlist that maximizes the reward.
The recommendation system suggests a playlist to a new user at each round, whose appreciation is revealed through multiple steps.
In particular, every partial observation corresponds to a song in the playlist, and the associated reward is positive if the user listens to that song and non-positive otherwise.
The cumulative reward provided by recommending a playlist to a single user corresponds to the sum of the reward terms from all the playlist songs.
Notice that the playlist cannot be trivially modeled as a collection of independent songs, as their order in the playlist affects the user's behavior.
In the classical delayed-feedback bandit setting, the feedback on the recommended playlist is obtained only once the user finishes listening to the entire playlist.
However, the platform monitors whether every song is listened to or skipped by the user.
Therefore, clues on the performances of the recommended arm can be exploited \emph{before} the user finishes the playlist.
}

\textcolor{black}{
Another scenario captured by the TP-MAB framework is the evaluation of medical treatments taking place over a long period of time.
In this setting, the per-round reward corresponds to the patient's state of health at each daily/weekly medical check, and the goal is to find the treatment providing the greatest overall benefit to the patient.
In the case of severe pathologies, such as cancer, this type of \emph{partial information} would span several months if not years, providing valuable insights that would be otherwise ignored. 
Applying a standard delayed-MAB approach to this scenario, \emph{i.e.}, taking decisions only at the end of each treatment cycle, could negatively affect the time required to select an effective medical treatment.
In this type of setting, we argue that the partial information provided by patients in periodic medical checks should be used to speed up the learning process. 
}

\paragraph{Original Contributions.}
\textcolor{black}{
Initially, we focus on the lower bound of TP-MABs, showing that the TP-MAB setting has the same regret lower bound of the standard delayed MAB setting when there is no further assumption about how the rewards are partitioned over time.
%
%
Since in many practical applications of interest the cumulative reward of each arm does not concentrate excessively in a short sub-range of rounds, we introduce a property describing how the maximum per-round reward distributes.
We call this property \emph{$\alpha$-smoothness} where $\alpha \geq 1$. 
In particular, the minimum value of $\alpha=1$ corresponds to the case in which there is no structure and, therefore, the maximum per-round reward can be the entire cumulative reward.
On the other hand, the maximum value of $\alpha$ is equal to the maximum delay and corresponds to the case in which the cumulative reward distributes evenly over time.
%
Thus, the maximum per-round reward decreases as the value of $\alpha$ increases. 
We show that the lower bound of this setting is of a factor $1 / \alpha$ smaller than that when $\alpha$-smoothness does not hold.
Then, we design two novel algorithms, namely \texttt{TP-UCB-FR} and \texttt{TP-UCB-EW}, suited for the TP-MAB setting, which exploit partial feedback and the $\alpha$-smoothness property.
We show that the regret of \texttt{TP-UCB-FR} is $\mathcal{O}(\ln T/\alpha)$, where $T$ is the time horizon of the learning process, and the regret of \texttt{TP-UCB-EW} is $\mathcal{O}(\ln T)$.
A comprehensive analysis the regret bounds of our and state-of-the-art algorithms  in various settings can be found in Table~3 (in Appendix~A for reasons of space).
Finally, we experimentally show that our algorithms outperform the state of the art over synthetically generated and a real-world playlist recommendation scenario.
%
}

\paragraph{Related Works.}
\textcolor{black}{
To the best of our knowledge, ours is the first work addressing a bandit problem in which the reward from a pull is partitioned across multiple rounds. 
The most related works concern the Delayed-MAB setting, such as the seminal paper by~\citeauthor{joulani2013online}~[\citeyear{joulani2013online}], which summarizes the known results on the regret upper bounds of online learning algorithms. 
%
%
They also provide a modification of the well-known UCB1 algorithm from~\citeauthor{auer2002finite}~[\citeyear{auer2002finite}] for the delayed-feedback setting, called Delayed-UCB1.
More recently, a variety of delayed-feedback scenarios were studied investigating directions different from ours, such as linear and contextual~(\citeauthor{arya2020randomized}~[\citeyear{arya2020randomized}], \citeauthor{vernade2020linear}~[\citeyear{vernade2020linear}], \citeauthor{zhou2019learning}~[\citeyear{zhou2019learning}]), non-stationary~(\citeauthor{vernade2020non}~[\citeyear{vernade2020non}]) bandits under delayed feedback. 
\citeauthor{pike2018bandits}~[\citeyear{pike2018bandits}] and~\citeauthor{cesa2018nonstochastic}~[\citeyear{cesa2018nonstochastic}] also analyze the case of delayed, aggregated, and anonymous feedback. 
For clarity, we remark that, in our work, per-round rewards corresponding to different pulls can be received in the same round, and it is known from which arm they were generated.
Many works apply bandits to practical scenarios, \emph{e.g.}, scheduling~\cite{cayci2019learning}, advertising~\cite{nuara2018combinatorial,castiglioni2022safe,nuara2022online}, pricing~\cite{trovo2018improving}, and delayed feedback settings~\cite{vernade2017stochastic}.
}

\textcolor{black}{
Works from the bandit literature, such as the ones by~\citeauthor{dudik2011efficient}~[\citeyear{dudik2011efficient}], \citeauthor{desautels2014parallelizing}~[\citeyear{desautels2014parallelizing}], \citeauthor{neu2013online}~[\citeyear{neu2013online}], rely on known constant delays or maximum delay values.  
Similarly, in our work, we assume a maximum finite delay equal to $\Tmax$, which is compliant with the real-world scenarios we aim at modeling, \emph{e.g.}, in the above example of playlist recommendations, an infinite $\Tmax$ would correspond to a playlist of an infinite number of songs.
According to the terminology used in the delayed-MAB literature, our setting is \emph{uncensored}, meaning that the reward provided by a given action is eventually observed after a finite maximum delay.
Conversely, many works in the field, such as, \emph{e.g.}, \citeauthor{gael2020stochastic}~[\citeyear{gael2020stochastic}] and \citeauthor{vernade2017stochastic}~[\citeyear{vernade2017stochastic}], deals with random delays from an unbounded distribution with finite expectation.
%
%
}

\section{Problem Formulation}

Consider a MAB problem with $K \in \mathbb{N}^*$ arms, over a time horizon of $T \in \mathbb{N}^*$ rounds.
At every round $t \in [T]$, the learner pulls an arm $i \in \mathcal{A} = [K]$ and, from the pull of that arm, gets a \emph{per-round reward} $x_{t,m-t+1}^i$ at every round $m \in \{t, \ldots, t + \Tmax - 1\}$, where $\Tmax \in \mathbb{N}^*$ is the time span over which the reward is partitioned.\footnote{We denote by $[n]$ the set $\{1, \ldots, n \}$}
In particular, $\Tmax - 1$ is the maximum delay affecting the observation of a per-round reward, whose value is known to the learner.
Therefore, at round $t + \Tmax - 1$, the cumulative reward from pulling arm $i$ at round $t$ is completely collected by the learner.
Furthermore, we denote by $\bm{x}^i_t = [x_{t,1}^i, \ldots, x_{t,\Tmax}^i]$ the vector of per-round rewards collected from pulling arm $i$ at round $t$.
For every $j \in [\Tmax]$, the per-round reward $x_{t,j}^i$ is a realization of a random variable $X_{t,j}^i$ with support $[\underline{X}_{j}^i, \overline{X}_{j}^i]$.
The cumulative reward collected from pulling arm $i$ at round $t$ is denoted by $r_{t}^i$, and it is the realization of the random variable $R_t^i := \sum_{j=1}^{\Tmax}{X_{t,j}^i}$, with support $[\underline{R}^i, \overline{R}^i]$, where $\underline{R}^i := \sum_{j=1}^{\Tmax}\underline{X}_{j}^i$, and $\overline{R}^i := \sum_{j=1}^{\Tmax}\overline{X}_{j}^i$.
For every $i \in \mathcal{A}$ and $t \in [T]$, we assume that the variables $R^i_t$ are independent with mean $\mu_i := \mathbb{E}[R^i_t]$.\footnote{
W.l.o.g., we assume $\underline{X}_{j}^i = 0, \forall i \in [K], \forall j \in [\Tmax]$.
}

A policy $\mathfrak{U}$ is an algorithm that at each round $t$ chooses an arm $i_t \in [K]$ .
The performance of a policy $\mathfrak{U}$ is evaluated in terms of \emph{pseudo-regret}, defined as the cumulative loss due to playing suboptimal arms during the time horizon $T$, formally:
\begin{equation*}
	\mathcal{R}_T(\mathfrak{U}) = T \mu^* - \mathbb{E} \left[\sum_{t=1}^{T} \mu_{i_t} \right],
\end{equation*}
where $\mu^* = \max_{i \in \mathcal{A}} \{\mu_i\}$ is the expected reward of the optimal arm $i^*$, and the expectation is taken w.r.t.~the stochasticity of the policy $\mathfrak{U}$.
Notice that we adopt the concept of pseudo-regret as for standard bandits, unlike what is done by~\citeauthor{vernade2017stochastic}~[\citeyear{vernade2017stochastic}], since our choice allows for a direct comparison with the vast prior work on delayed bandits.

In what follows, we cast the playlist recommendation problem, described in the introduction, in the TP-MAB setting.
\begin{example}[Playlist Recommendation] \label{ex:rec}
	At each round $t$, a new user enters the platform, which provides a playlist suggestion.
	The different arms $i$ are the available playlists to suggest, each composed of $N$ songs.
	Songs are characterized by $4$ listening levels (from ``skipped'' to ``complete''), each associated with a different Bernoulli random variable representing the corresponding per-round reward.
	The vector of realized per-round rewards of song $k \in [N]$ is $[x^i_{t,4(k-1)+1}, x^i_{t,4(k-1)+2}, x^i_{t,4(k-1)+3}, x^i_{t,4(k-1)+4}]$.
	Each variable assumes a value of $1$ if the user reaches the corresponding level, and a value of $0$ if the user stops listening to the song before that level.
	The cumulative reward $R^i_t$ for pulling arm $i$ at round $t$ is the sum of the rewards from the songs in the playlist, and the time span over which the platform observes the reward is $\Tmax = 4 N$.
\end{example}

We show that the TP-MAB problem has a lower-bound on the regret of the same order of the delayed-feedback bandit problem.
The rationale is that no better lower bound is possible as delayed-feedback MABs with a finite delay are a subclass of TP-MABs whose reward vector $\bm{x}^i_t$ has a single non-zero element for each $i \in \mathcal{A}$ and $t \in [T]$.
Most interestingly, the worst-case instance for the regret lower bound in the TP-MAB setting is the delayed-feedback bandit.\footnote{
All the proofs are deferred to Appendix~B for space reasons. See \url{https://trovo.faculty.polimi.it/01papers/romano2022multi.pdf}.}
\begin{restatable}{theorem}{thmlower} \label{thm:lower}
	The regret of any uniformly efficient policy $\mathfrak{U}$ applied to the TP-MAB problem is bounded from below by:
\begin{equation}
	\lim \inf_{T \rightarrow +\infty} \frac{\mathcal{R}_T(\mathfrak{U})}{\ln T} \geq \sum_{i:\mu_i < \mu^*} \frac{\Delta_i}{KL\left( \frac{\mu_i}{\overline{R}_{\max}}, \frac{\mu^*}{\overline{R}_{\max}} \right)},
\end{equation}
where $\Delta_i := \mu^* - \mu_i$ is the expected loss suffered by the learner if the arm $i$ is chosen instead of the optimal one $i^*$, $\overline{R}_{\max} := \max_{i \in [K]} \overline{R}^i$, and $KL(p,q)$ is the Kullback-Leibler divergence between Bernoulli r.v.~with means $p$ and $q$.\footnote{
An uniformly efficient policy chooses the suboptimal arms on average $o(t^a)$ times ($0 < a < 1$) over $t$ rounds.}
\end{restatable}
Notice that the lower bound holds for general TP-MAB problems.
In the following section, we show that focusing on a broad subset of instances of practical interest, we can design algorithms with a better regret upper bound.

\section{$\alpha$-Smoothness Property} \label{S:r_struct}
From Theorem~\ref{thm:lower}, we know that we cannot design algorithms with regret upper bounds better than those of the algorithms for the delayed-feedback bandit setting.
Nonetheless, in practice, collecting per-round rewards can provide useful information on the cumulative reward of an arm.
However, as already pointed out by~\citeauthor{gael2020stochastic}~[\citeyear{gael2020stochastic}] for the standard delayed-feedback setting, zero rewards are ambiguous since they do not give any information on future rewards.
In the general setting, small per-round rewards observed in the first rounds after the pull are not much informative to bound the values of future ones.
To avoid this, we focus on those problems in which the maximum reward realized over a few rounds cannot exceed a fraction of the maximum reward $\overline{R}^i$.

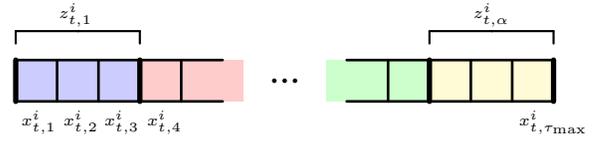
\begin{figure}[t!]
	\centering
	\vspace{-0.8cm}
	\begin{tikzpicture}[line cap=round,line join=round,>=triangle 45,x=1cm,y=1cm, scale=0.55, font=\tiny ]
\clip(5.5,7) rectangle (20,12);
\draw[fill=blue!20,draw=none] (6,8) rectangle (9,9);
\draw[fill=red!20,draw=none] (9,8) rectangle (11.5,9);
\draw[fill=green!20,draw=none] (13.5,8) rectangle (16,9);
\draw[fill=yellow!20,draw=none] (16,8) rectangle (19,9);
\draw [line width=2pt] (6,9)-- (6,8);
\draw [line width=1pt] (6,9)-- (9,9);
\draw [line width=2pt] (9,9)-- (9,8);
\draw [line width=1pt] (6,8)-- (7,8);
\draw [line width=1pt] (7,8)-- (8,8);
\draw [line width=1pt] (8,8)-- (9,8);
\draw [line width=1pt] (7,9)-- (7,8);
\draw [line width=1pt] (8,9)-- (8,8);
\draw [line width=1pt] (9,9)-- (10,9);
\draw [line width=1pt] (9,8)-- (10,8);
\draw [line width=1pt] (10,9)-- (10,8);
\draw [line width=1pt] (10,8)-- (11,8);
\draw [line width=1pt] (10,9)-- (11,9);
\draw [line width=1pt] (14,9)-- (15,9);
\draw [line width=1pt] (15,9)-- (15,8);
\draw [line width=1pt] (15,8)-- (16,8);
\draw [line width=1pt] (15,9)-- (16,9);
\draw [line width=2pt] (16,9)-- (16,8);
\draw [line width=1pt] (16,9)-- (17,9);
\draw [line width=1pt] (17,9)-- (18,9);
\draw [line width=1pt] (18,9)-- (19,9);
\draw [line width=2pt] (19,9)-- (19,8);
\draw [line width=1pt] (17,9)-- (17,8);
\draw [line width=1pt] (18,9)-- (18,8);
\draw [line width=1pt] (14,8)-- (15,8);
\draw [line width=1pt] (16,8)-- (17,8);
\draw [line width=1pt] (17,8)-- (18,8);
\draw [line width=1pt] (18,8)-- (19,8);
\draw [line width=0.8pt] (6,9.4)-- (6,9.7);
\draw [line width=0.8pt] (6,9.7)-- (9,9.7);
\draw [line width=0.8pt] (9,9.7)-- (9,9.4);
\draw [line width=0.8pt] (7.5,9.7)-- (7.5,9.7);
\draw [line width=0.8pt] (16,9.4)-- (16,9.7);
\draw [line width=0.8pt] (16,9.7)-- (19,9.7);
\draw [line width=0.8pt] (19,9.7)-- (19,9.4);
\draw [line width=0.8pt] (17.5,9.7)-- (17.5,9.7);
\begin{scriptsize}
\draw[color=black] (6.56,7.5) node {$x_{t,1}^i$};
\draw[color=black] (7.58,7.5) node {$x_{t,2}^i$};
\draw[color=black] (8.565,7.5) node {$x_{t,3}^i$};
\draw[color=black] (9.6,7.5) node {$x_{t,4}^i$};
\draw [fill=black] (12.75,8.5) circle (1.2pt);
\draw [fill=black] (12.25,8.5) circle (1.2pt);
\draw [fill=black] (12.5,8.5) circle (1.2pt);
\draw[color=black] (19,7.5) node {$x_{t,\Tmax}^i$};
\draw[color=black] (7.45,10.1) node {$z_{t,1}^i$};
\draw[color=black] (17.5,10.1) node {$z_{t,\alpha}^i$};
\end{scriptsize}
\end{tikzpicture}
	\caption{Example of $\alpha$-smooth reward with $\step = 3$.}
	\label{fig:ex_alpha_smoothness}
	\vspace{-0.3cm}
\end{figure}

Let us consider $\alpha \in [\Tmax]$ s.t.~$\alpha$ is a factor of $\Tmax$, \emph{i.e.}, $\frac{\Tmax}{\alpha} =: \step$ and $\step \in \mathbb{N}$.\footnote{
We assume $\alpha$ is a factor of $\Tmax$ for the sake of presentation.
The following results also hold for generic $\alpha \in [\Tmax]$.}
Let us define the vector $\bm{Z}^i_{t,\alpha} := \left[Z^i_{t,1}, \ldots , Z^i_{t,\alpha} \right]$ whose element $Z^i_{t,k}$ is the random variable corresponding to the sum of a set of consecutive per-round rewards of cardinality $\step$.
Formally, for every $k \in [\alpha]$:
\begin{equation}
	Z^i_{t,k} := \sum_{j=(k-1)\step + 1}^{k \step} X^i_{t,j}.
\end{equation}
The support of $Z^i_{t,k}$ is denoted by $[\underline{Z}^i_{\alpha,k}, \overline{Z}^i_{\alpha,k}]$,
where $\underline{Z}^i_{\alpha,k} := \sum_{j=(k-1)\step + 1}^{k \step} \underline{X}_{j}^i$, and $\overline{Z}^i_{\alpha,k} := \sum_{j=(k-1)\step + 1}^{k \step} \overline{X}_{j}^i$.
Intuitively, the $\alpha$-smoothness property states that the elements in $\bm{Z}^i_{t,\alpha}$ are independent and that, when $\alpha > 1$, the maximum reward $\overline{R}^i$ of a pull cannot be realized in a single time span corresponding to a $Z^i_{t,k}$ element.
Formally:
\begin{definition}[$\alpha$-smoothness]  \label{def:asmooth}
	In the TP-MAB setting, for $\alpha \in [\Tmax]$, we say that the reward is \emph{$\alpha$-smooth} if and only if $\frac{\Tmax}{\alpha} = \step$, with $\step \in \mathbb{N}$, and, for each $k \in [\alpha]$, the random variables $Z^i_{t,k}$ are independent and s.t.~$\overline{Z}^i_{\alpha,k}=\overline{Z}^i_{\alpha} = \frac{\overline{R}^i}{\alpha}$.
\end{definition}
An example of $\alpha$-smooth environment with $\phi = 3$ is presented in Figure~\ref{fig:ex_alpha_smoothness}, where colors denote the elements $z^i_{t,k}$ that are the realizations of the variables $Z^i_{t,k}$.

Consider the extreme values of parameter $\alpha$.
When $\alpha = 1$, the reward has no constraint on how it distributes over time.
This scenario includes the delayed-feedback bandit setting in which the cumulative reward provided by the arm pulled at $t$ is entirely collected at a single round (including the last possible round $t + \Tmax - 1$).
Note that, in this case, at each round before $t + \Tmax - 1$, the sum of the future per-round rewards is in the range $[0, \overline{R}^i]$.
Conversely, when $\alpha = \Tmax$, the vector of aggregated rewards coincides with the vector of per-round rewards, \emph{i.e.}, $\bm{Z}^i_{t,\Tmax} = \bm{X}^i_t$, and each per-round reward is at most $\overline{X}^i_j = \overline{R}^i / \Tmax$.
Thus, observing low rewards in the first rounds after the pull provides useful information on the actual cumulative reward.
In particular, after observing the first $n < \Tmax$ per-round rewards, we know that the cumulative reward achievable in the following rounds is in the range $[0, \frac{\Tmax-n}{\Tmax} \overline{R}^i]$.
This information dramatically reduces the uncertainty on the future rewards w.r.t.~a setting without smooth rewards (\emph{e.g.}, $\alpha = 1$).
The $\alpha$-smoothness property characterizes those setting where not gaining much in the first rounds precludes the possibility of achieving the maximum possible reward over the entire interval.

Consider the playlist recommendation problem in Example~\ref{ex:rec}.
Since the reward corresponding to a song is composed of $4$ Bernoulli variables and has a maximum of $\overline{Z}^i_{\alpha} = 4$, $\alpha$-smoothness holds with $\alpha = \frac{\overline{R}^i}{\overline{Z}^i_{\alpha}} = \frac{4N}{4} = N$.

Assuming $\alpha$-smoothness, we have a lower bound of:
\begin{restatable}{theorem}{thmloweralpha} \label{thm:loweralpha}
The regret of any uniformly efficient policy $\mathfrak{U}$ applied to the TP-MAB problem with the $\alpha$-smoothness property is bounded from below by:
\begin{equation}
	\lim \inf_{T \rightarrow +\infty} \frac{\mathcal{R}_T(\mathfrak{U})}{\ln T} \geq \sum_{i:\mu_i < \mu^*} \frac{\Delta_i}{\alpha KL\left( \frac{ \mu_i}{\overline{R}_{\max}}, \frac{\mu^*}{\overline{R}_{\max}} \right)}.
\end{equation}
\end{restatable}
We remark that this bound is tighter than the one provided in Theorem~\ref{thm:lower} by a multiplicative factor of $1/\alpha$.

\section{Algorithms for the TP-MAB Setting} \label{s:new_algo}

We propose two novel algorithms, namely Temporally-Partitioned rewards UCB with Fictitious Realizations (\texttt{TP-UCB-FR}) and Temporally-Partitioned rewards Element-Wise UCB (\texttt{TP-UCB-EW}), for the TP-MAB problem, which aim at maximizing the cumulative reward and exploit the $\alpha$-smoothness property to do that.
From now on, we denote the two corresponding policies by $\mathfrak{U}_{\texttt{FR}}$ and $\mathfrak{U}_{\texttt{EW}}$, respectively.
%
%

\subsection{The \texttt{TP-UCB-FR} Algorithm}

\begin{algorithm}[t!]
	\caption{\texttt{TP-UCB-FR}} \label{a:pd-ucb1-fr}
	\begin{algorithmic}[1]
		\State $\textbf{Input}$: $\alpha \in [\Tmax]$, $\Tmax \in \mathbb{N}^*$
		\For { $t \in  \{1, \ldots, K \}$} \Comment{\textcolor{gray}{init phase}}
		\State Pull arm $i_t = t$ \label{line:init1}
		\EndFor
		\For { $t \in  \{K+1, \ldots, T \}$} \Comment{\textcolor{gray}{loop phase}}
			\For { $i \in  \{1, \ldots, K \}$}
				\State Compute $\hat{R}^{i}_{t-1}$ and $c^i_{t-1}$ as in Eq.s~\eqref{eq:exprew1}-\eqref{eq:bound1}
				\State $u^i_{t-1} \gets \hat{R}^{i}_{t-1} + c^i_{t-1} $
			\EndFor
		\State Pull arm $i_t = \arg \max_{i \in [K]}$ $u^i_{t-1}$ \label{line:upper1}
		\State Observe $x^{i_h}_{h,t-h+1}$ for  $h \in \{t-\Tmax + 1, \ldots, t\}$
		\label{line:obs1}
		\EndFor
	\end{algorithmic}
\end{algorithm}

The pseudo-code of \texttt{TP-UCB-FR} is provided in Algorithm~\ref{a:pd-ucb1-fr}.
The rationale is to use the rewards coming from not fully-realized reward vectors by replacing the missing elements with fictitious realizations.
At round $t$, fictitious reward vectors are associated to each arm pulled in the time span $H:=\{t-\Tmax+1, \ldots, t-1\}$.
We denote them by $\tilde{\bm{x}}^i_h = [\tilde{x}^i_{h,1}, \ldots, \tilde{x}^i_{h,\Tmax}]$ with $h \in H$, where $\tilde{x}^i_{h,j} := x^i_{h,j}$, if $h + j \leq t$, and $\tilde{x}^i_{h,j} = 0$, if $h + j > t$.
%
%
The corresponding fictitious cumulative reward is $\tilde{r}^i_h := \sum_{j=1}^{\Tmax}\tilde{x}^i_{h,j}$.
The algorithm takes as input the smoothness $\alpha \in [\Tmax]$, and the maximum delay $\Tmax$.\footnote{
If these information are not available one should use $\alpha = 1$, meaning we are not assuming any structure over the reward, and use as $\Tmax$ the largest delay observed so far.}
During the initialization phase, all arms are pulled once (Line~\ref{line:init1}).
After that, at each round $t$, it computes the estimated expected reward for each arm $i$:
\begin{align} \label{eq:exprew1}
	\hat{R}^{i}_{t-1} := \frac{1}{n^i_{t-1}} \left( \sum_{h=1}^{t-\Tmax} \hspace{-0.2cm} r^i_h \mathds{1}_{\{i_h = i\}} + \sum_{h \in H} \tilde{r}^i_h \mathds{1}_{\{i_h = i\}} \right),
\end{align}
where $n^i_{t-1} := \sum_{h=1}^{t-1} \mathds{1}_{\{i_h = i\}}$ is the number of times arm~$i$ has been pulled by the policy up to round $t-1$, and the confidence interval:
\begin{equation} \label{eq:bound1}
	c^i_{t-1} := \overline{R}^i \sqrt{\frac{2 \ln (t-1)}{\alpha n^i_{t-1}}} + \frac{\step(\alpha+1) \overline{R}^i}{2 n^i_{t-1}}.
\end{equation}
Finally, it pulls the arm with the largest upper confidence bound $u^i_{t-1}$ (Line~\ref{line:upper1}), and observes its reward (Line~\ref{line:obs1}).

We provide the following upper bound on the regret:
\begin{restatable}{theorem}{FRreg} \label{t:FRreg}
	In the TP-MAB setting with $\alpha$-smooth reward, the pseudo-regret of \emph{\texttt{TP-UCB-FR}} after $T$ rounds is:
	\begin{align*}
	   \mathcal{R}_T(\mathfrak{U}_{\emph{\texttt{FR}}})  \leq &\sum_{i:\mu_i<\mu^*} \frac{4(\overline{R}^i)^2 \ln T}{\alpha \Delta_i} \mleft( 1 + \sqrt{1 + \frac{\alpha(\alpha + 1) \phi \Delta_i}{2 \overline{R}^i \ln T}} \mright) \\
	&+  (\alpha+1) \step \sum_{i:\mu_i<\mu^*} \overline{R}^i + \mleft( 1 + \frac{\pi^2}{3} \mright) \sum_{i : \mu_i < \mu^*} \Delta_i.
	\end{align*}
\end{restatable}
We observe that the dominant term in $T$ has the order of $\mathcal{O} \left( \sum_{i:\mu_i < \mu^*} \frac{ \overline{R}_{\max}^2 \ln T}{\alpha \Delta_i} \right)$, where $\overline{R}_{\max} = \max_i \overline{R}^i$.
When $\alpha = 1$, the upper bound scales as the one of classical MAB algorithms in stochastic settings.
Notice that the pseudo-regret indirectly depends on $\Tmax$ since $\overline{R}^i$ represents the cumulative reward obtained over $\Tmax$ rounds.
Let us compare this result with the one provided in Theorem~\ref{thm:lower} for general TP-MAB problems.
Applying to Theorem~\ref{thm:lower} the inequality
$KL(p,q) \leq \frac{(p-q)^2}{q(1-q)}$,
where for $p, q \in [0,1]$, derived using the fact that $\ln x \leq x - 1$, we get:
\begin{equation}
	\lim \inf_{T \rightarrow +\infty} \frac{\mathcal{R}_T(\mathfrak{U})}{\ln T} \geq \sum_{i:\mu_i < \mu^*} \frac{\beta}{\Delta_i},
\end{equation}
where $\beta = \frac{\mu^*}{\overline{R}_{\max}} \mleft(1 - \frac{\mu^*}{\overline{R}_{\max}}\mright)$.

For $\alpha > 4(\overline{R}^i)^2/\beta$, the multiplicative factor in the dominant term of the upper bound provided in Theorem~\ref{t:FRreg} is better than that in the lower bound in Theorem~\ref{thm:lower}.
This suggests that exploiting the $\alpha$-smoothness provides an improvement over the classical and delayed-feedback MABs.

\subsection{The \texttt{TP-UCB-EW} Algorithm}

\begin{algorithm}[t!]
	\caption{\texttt{TP-UCB-EW}} \label{a:pd-ucb1-ew}
	\begin{algorithmic}[1]		
		\State $\textbf{Input}$: $\alpha \in [\Tmax]$, $\Tmax \in \mathbb{N}^*$
		\For { $t \in  \{1, \ldots, K \}$} \Comment{\textcolor{gray}{init phase}}
			\State Pull arm $i_t = t$ \label{line:init2}
		\EndFor
		\For { $t \in  \{K+1, \ldots, T\}$} \Comment{\textcolor{gray}{loop phase}}
			\For { $i \in  \{1, \ldots, K\}$}
				\For { $k \in  \{1, \ldots, \alpha\}$}
				    \State Compute $\hat{Z}^{i}_{t-1, k}$ and $c^i_{t-1, k}$ as in Eq.s~\eqref{eq:exprew2}-\eqref{eq:bound2}
				\EndFor
			\State $u^i_{t-1} \gets \sum_{k=1}^{\alpha} \mleft(\hat{Z}^{i}_{t-1,k} + c^i_{t-1,k} \mright) $ \label{line:ucbi}
			\EndFor
		\State Pull arm $i_t \in \arg \max_{i \in [K]}$ $u^i_{t-1}$ \label{line:pull}
		\State Observe $x^{i_h}_{h,t-h+1}$ for $h \in \{t-\Tmax + 1, \ldots, t\}$ \label{line:obs2}		
		\EndFor	
	\end{algorithmic}
\end{algorithm}

The pseudo-code of \texttt{TP-UCB-EW} is provided in Algorithm~\ref{a:pd-ucb1-ew}.
The key idea is to compute an upper confidence bound for the average of each set of $k$-th realized aggregated rewards $z^i_{t,k}$ from arm $i$ and use them to build an upper bound on the overall average reward $R^i_t$.
It takes as input the smoothness parameter $\alpha$, and the maximum delay parameter $\Tmax$.
At first, it pulls each arm once (Line~\ref{line:init2}), while, in the following rounds, it computes the empirical mean:
\begin{equation} \label{eq:exprew2}
    \hat{Z}^{i}_{t-1,k} := \frac{\sum_{h=1}^{t - k \step  } z^{i}_{h,k} \mathds{1}_{\{i_h = i\}}}{n^i_{t-1,k}},
\end{equation}
where $n^i_{t-1,k} := \sum_{h=1}^{t - k \step} \mathds{1}_{\{i_h = i\}}$ is the cardinality of the rewards observed up to round $t-1$ for the $k$-th element of $\bm{Z}^i_{t-1,\alpha}$, and the confidence bound:
\begin{equation} \label{eq:bound2}
    c^i_{t-1,k} := \frac{\overline{R}^i}{\alpha}\sqrt{\frac{2 \ln (t-1)}{n^i_{t-1,k}}}.
\end{equation}
We remark that $\hat{Z}^{i}_{t-1, k} + c^i_{t-1, k}$ is an upper confidence bound for the $k$-th element of $\bm{Z}^i_{t-1,\alpha}$.
Finally, the algorithm computes the upper bound $u^i_{t-1}$, summing the bounds above (Line~\ref{line:ucbi}), selects the arm $i$ choosing the largest $u^i_{t-1}$ (Line~\ref{line:pull}), and observes its reward (Line~\ref{line:obs2}).

We provide the following upper bound on the regret:
\begin{restatable}{theorem}{EWreg}\label{t:EWreg}
	In the TP-MAB setting with $\alpha$-smooth reward, the pseudo-regret of \emph{\texttt{TP-UCB-EW}} after $T$ rounds is:
	\begin{align*}
		\R_T(\mathfrak{U}_{\emph{\texttt{EW}}}) & \leq \hspace{-0.3cm} \sum_{i:\mu_i<\mu^*}
	\frac{8(\overline{R}^i)^2\ln T}{\Delta_i} \hspace{-0.05cm} + \hspace{-0.05cm} \alpha\mleft(\step +\frac{\pi^2}{3}\mright) \hspace{-0.2cm} \sum_{i : \mu_i < \mu^*} \hspace{-0.3cm} \Delta_i.
	\end{align*}
\end{restatable}
Focusing on the dominant term in $T$ of the regret bound, we do not have an explicit improvement over the classical and delayed-feedback MAB algorithms.
Therefore, in this case, the structure provided by the $\alpha$-smoothness seems not to affect the regret bound.
Hence, from an asymptotic point of view, there is not a clear advantage from having $\alpha$-smooth rewards.
However, the constant term is significantly smaller than that of \texttt{TP-UCB-FR} and allows \texttt{TP-UCB-EW} to be much more effective than \texttt{TP-UCB-FR} to tackle TP-MAB problems with a short time horizon.

\section{Empirical Evaluation} \label{S:exp}

We compare \texttt{TP-UCB-FR} and \texttt{TP-UCB-EW} algorithms with the \texttt{UCB1} algorithm by~\citeauthor{auer2002finite}~[\citeyear{auer2002finite}] and the \texttt{Delayed-UCB1} algorithm by~\citeauthor{joulani2013online}~[\citeyear{joulani2013online}] in $\alpha$-smooth TP-MAB environments.
Appendix~A provides details on the adaptation of these two  state-of-the-art algorithms to the TP-MAB problem.
Notice that, for \texttt{UCB1}, we assume to immediately get the cumulative reward of a pull. Therefore, it represents a \emph{clairvoyant} algorithm observing $R^i_t$ at round $t$.
%
%
%
We compare the algorithms in three settings: two synthetically-generated environments and a real-world playlist recommendation scenario.\footnote{
More details about the experiments are deferred to Appendix~C.}

\vspace{0.15cm}
\paragraph{Setting \#1.}
At first, we evaluate the influence of the parameter $\alpha$.
We model $K = 10$ arms, whose maximum reward is s.t.~$\overline{R}^i = 100 i$.
The reward is collected over $\Tmax = 100$ rounds, the smoothness parameter is $\alpha = 20$, and the aggregated rewards are s.t.~$Z^i_{t,k} \sim \frac{\overline{R}^i}{\alpha} \textnormal{U}([0, 1])$, for each $k \in [\alpha]$.
We run the algorithms over a time horizon of $T = 10^5$ and average the results over $50$ independent runs.
%
%
In the results, \texttt{TP-UCB-FR}($\eta$) and \texttt{TP-UCB-EW}($\eta$) are s.t.~the value of $\alpha$ taken as input is $\eta$, with $\eta \in \{5, 10, 20, 25, 50\}$.

\begin{figure}[t!]
	\centering
	\input{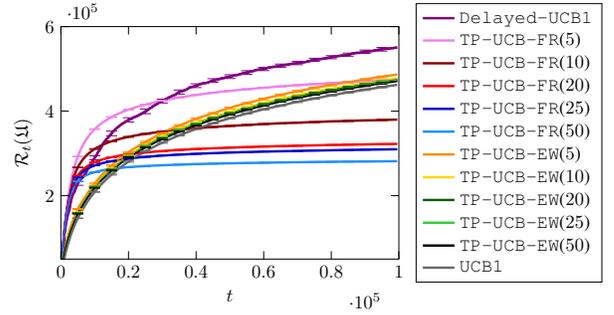}
	\caption{Pseudo-regret over time for Experimental Setting \#1.} \label{f:alpha}
\end{figure}

\vspace{0.15cm}
\noindent \emph{Results.}
Figure~\ref{f:alpha} shows the pseudo-regret $\mathcal{R}_t(\mathfrak{U})$ over the time horizon and the vertical bars represent the $95\%$ confidence intervals for the mean value. 
Let us focus on \texttt{TP-UCB-FR}($20$) and \texttt{TP-UCB-EW}($20$), for which $\eta$ is equal to the $\alpha$ of the environment.
%
\texttt{TP-UCB-EW}($20$) provides better results than \texttt{Delayed-UCB1} over the entire time horizon, while \texttt{TP-UCB-FR}($20$) is better than \texttt{Delayed-UCB1} for $t > 10^4$ and better than \texttt{TP-UCB-EW}($20$) for $t > 2 \cdot 10^4$.
This suggests that \texttt{TP-UCB-FR}($20$) is more suitable for longer time horizons, and this behavior is confirmed by the asymptotic order of Theorem~\ref{t:FRreg}.
Notice that \texttt{UCB1} obtains the reward as soon as an arm has been pulled, but it does not exploit the $\alpha$-smoothness property.
\emph{Vice versa}, our algorithms incorporates this information that, in some specific situations, allows us to beat even the non-delayed baseline.
%
%
%

During rounds $t \in [1, 7000]$, the \texttt{Delayed-UCB1} algorithm outperforms \texttt{TP-UCB-FR}, since, during the initial rounds, incomplete samples may be far different from the corresponding unseen realizations, and, therefore, \texttt{TP-UCB-FR} initially pulls the suboptimal arms more often than \texttt{Delayed-UCB1}.
Nonetheless, \texttt{TP-UCB-FR} outperforms \texttt{Delayed-UCB1} over longer time horizons, as expected given the result in Theorem~\ref{t:FRreg}.
\texttt{TP-UCB-EW} has a similar asymptotic behavior of those of \texttt{UCB1} and \texttt{Delayed-UCB1}, \emph{i.e.}, the regret curves becomes parallel after $\approx 4000$ rounds.
This is because the overall exploration term of the three algorithms is of the same order in~$t$ and $\alpha$, and therefore the advantages of \texttt{TP-UCB-EW} are mainly experienced in the early stages of the learning process.
Summarily, for short-time horizons, \texttt{TP-UCB-EW} is preferable to~\texttt{TP-UCB-FR}, while \texttt{TP-UCB-FR} shows better performance over long periods.

Let us focus on the results obtained with \texttt{TP-UCB-FR}($\eta$).
Setting $\eta < \alpha$, \emph{i.e.}, underestimating the value of $\alpha$, provides worse results in terms of regret, while $\eta > \alpha$ seems to improve the performance of the algorithm without compromising the convergence properties.
This suggests that if the $\alpha$ parameter is unknown, one should use an optimistic (large) value in the algorithm.
Notice that the regret varies of $\approx 40\%$ w.r.t.~the different versions of \texttt{TP-UCB-FR} changing the value of $\eta$, which suggests that \texttt{TP-UCB-FR} is strongly influenced by a mis-specification of the parameter $\eta$.
Focusing on \texttt{TP-UCB-EW}($\eta$), we have a behaviour similar to the one observed for \texttt{TP-UCB-FR}($\eta$), showing how larger values for $\eta$ provide better results.
Conversely, the performance of \texttt{TP-UCB-EW} present a lower variability by changing the parameter $\eta$, and the gap in terms of regret among the different versions of \texttt{TP-UCB-EW} is of $\approx 3\%$.

\begin{table}[t!]
	\centering
	\renewcommand{\arraystretch}{1.1}
	\resizebox{0.8\columnwidth}{!}{
	\begin{tabular}{|c|c|c|c|}
		\hline
		$\Tmax$ & $\alpha$ & $\mathcal{R}_T^{(\%)}(\mathfrak{U}_{\texttt{FR}})$ & $\mathcal{R}_T^{(\%)}(\mathfrak{U}_{\texttt{EW}})$ \\
		\hline
		100 & 10  & 68.06$\%$ (0.26$\%$) & 86.03$\%$ (0.59$\%$) \\
		\hline
		200 & 20  & 95.42$\%$ (0.15$\%$) & 80.38$\%$ (0.34$\%$)\\
		\hline
		100 & 50  & 50.84$\%$ (0.11$\%$) & 85.36$\%$ (0.33$\%$) \\
		\hline
		200 & 100 & 81.55$\%$ (0.10$\%$) & 78.70$\%$ (0.24$\%$) \\
		\hline
	\end{tabular}}
	\caption{$\mathcal{R}_T^{(\%)}(\mathfrak{U})$ for Experimental Setting \#2.} \label{t:comment_exp2}
\end{table}

\vspace{0.15cm}
\paragraph{Setting \#2.}
We study the behavior of our algorithms in settings with different maximum delay $\Tmax$ and smoothness $\alpha$.
The scenario is the same presented in Setting \#$1$ except that the maximum reward for the arm $i$ is $\overline{R}^i = \Tmax \cdot i$.\footnote{
In Appendix~C, we also report experiments in scenarios differing in how the aggregated rewards are distributed over the $\phi$ elements composing $Z^i_{t,k}$, which confirm what is shown in this section.
}
%
%
We evaluate the algorithms in terms of percentage of the regret w.r.t.~the one provided by \texttt{Delayed-UCB1}, whose policy is denoted by $\mathfrak{U}_{\texttt{D}}$, formally $\mathcal{R}_T^{(\%)}(\mathfrak{U}) := \mathcal{R}_T(\mathfrak{U}) / \mathcal{R}_T(\mathfrak{U}_{\texttt{D}}) \cdot 100$.
We average the results over $50$ independent experiments.

\vspace{0.15cm}
\noindent \emph{Results.} 
Table~\ref{t:comment_exp2} provides the values of $\mathcal{R}_T^{(\%)}(\mathfrak{U})$ for our algorithms ($95\%$ CI in brackets).
In all the scenarios, the proposed algorithms outperform the \texttt{Delayed-UCB1} algorithm, providing a regret smaller than $95.5\%$ of the \texttt{Delayed-UCB1} one.
%
%
Comparing the results with the same maximum delay $\Tmax$ we notice that a larger value for $\alpha$ provides better performance.
This was expected since larger values for $\alpha$ imply that the \texttt{TP-UCB-FR} and \texttt{TP-UCB-EW} algorithms can better exploit the reward structure.
By comparing the settings with maximum delay $\Tmax = 100$ and $\Tmax = 200$, the two algorithms behave in opposite ways: the performance of \texttt{TP-UCB-EW} improves by more than $6\%$, while the regret of \texttt{TP-UCB-FR} increases of more than $30\%$.
This is due to the fact that, with larger $\Tmax$, \texttt{TP-UCB-FR} shows its better behaviour for larger time horizons.
%

\vspace{0.15cm}
\paragraph{Spotify Setting.}
We apply the TP-MAB approach to solve the user recommendation problem presented in Example~\ref{ex:rec}, using a dataset by Spotify~\cite{brost2019music}.
%
%
We select the $K = 6$ most played playlist as the arms to be recommended, and each time a playlist $i$ is selected, the corresponding reward realizations $\bm{x}^i_t$ for the first $N = 20$ songs is sampled from the listening sessions of that playlist contained in the dataset.
We recall that, in this setting, the maximum delay is $\Tmax = 4N = 80$, and the smoothness parameter is $\alpha = 20$.
More details on the setting and the distributions of the reward for each playlist are provided in Appendix~C.
We average the results over $50$ independent runs.

\begin{table}[t!]
	\centering
	\renewcommand{\arraystretch}{1.1}
	\resizebox{0.5\columnwidth}{!}{
	\begin{tabular}{|c|c|}
		\hline
		& $\mathcal{R}_T(\mathfrak{U})$  \\
		\hline
		\texttt{Delayed-UCB1} & 56473 (805) \\
		\hline
		\texttt{TP-UCB-FR} & 25367 (369)	 \\
		\hline
		\texttt{TP-UCB-EW} & 55000 (951) \\
		\hline
		\texttt{UCB1} & 47368	(1289) \\
		\hline
	\end{tabular}}
	\caption{Pseudo-regret for the Spotify experimental setting.} \label{t:comment_exp3}
\end{table}

\begin{figure}[t!]
	\centering
	\input{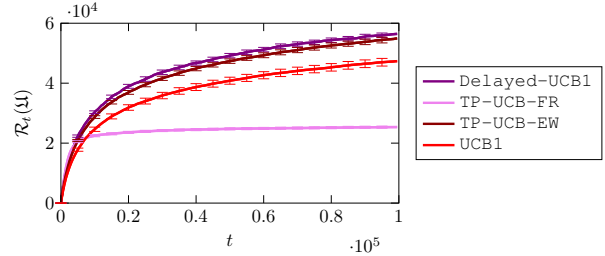}
	\caption{Pseudo-regret over time for the Spotify setting.} \label{fig:spotify}
	\vspace{-0.4cm}
\end{figure}

\vspace{0.15cm}
\noindent \emph{Results.}
Table~\ref{t:comment_exp3} shows that the \texttt{TP-UCB-FR} algorithm provides the best performance among the analysed algorithms, outperforming \texttt{UCB1} thanks to the exploitation of the $\alpha$-smoothness property.
%
%
%
The regret over time in Figure~\ref{fig:spotify} shows that the \texttt{TP-UCB-FR} provides worse performance than \texttt{TP-UCB-EW} only for a limited amount of rounds ($t < 4000$).
This suggests that, in this specific scenario, the \texttt{TP-UCB-FR} algorithm represents a good candidate to provide playlist recommendations.

\section{Conclusion and Future Works} \label{S:conclusion}

This paper introduces the novel TP-MAB setting, which generalizes the delayed-feedback bandit setting with bounded delay.
First, we show that the lower bound of the TP-MAB problem is the same of that of the standard delayed MAB problem.
Then, we characterize a broad set of reward structures, by defining the $\alpha$-smoothness property, for which we provide a tighter lower bound.
We design the \texttt{TP-UCB-FR} and the \texttt{TP-UCB-EW} algorithms, suited for the TP-MAB setting, which exploit the partial rewards collected over time and the $\alpha$-smoothness property.
We show that the upper bounds on the regret for these algorithms are $\mathcal{O}(\ln T/\alpha)$ and $\mathcal{O}(\ln T)$, respectively.
Finally, we empirically show that our algorithms outperforms the state of the art over a wide range of settings generated from synthetic and real-world data.
%

%
An interesting future extension would be to consider generic functions regulating the relationship between the cumulative and delayed rewards.
%
%
%


\clearpage
\bibliographystyle{named}
\bibliography{ijcai22}

\clearpage
\onecolumn
\appendix
%

\begin{center}
	\LARGE{Appendix of the Paper ``Multi-Armed Bandit Problem with Temporally-Partitioned Rewards: When Partial Feedback Counts''}
\end{center}

\begin{algorithm}[h]
	\caption{\texttt{UCB1}}
	\begin{algorithmic}[1]	
		\For { $t \in  \{1, \ldots, K\}$} \Comment{init phase}
		\State Pull arm $i_t = t$ \label{line:initucb}
		\State Observe the reward $r^{i_t}_t$ of the arm pulled at round $t$
		\EndFor
		\For { $t \in  \{K+1, \ldots, T\}$} \Comment{loop phase}
		\For { $i \in  \{1, \ldots, K\}$}
		\State $\hat{R}^{i}_{t-1} \gets \frac{1}{n^i_{t-1}} \sum_{h=1}^{t-1}r^i_h \mathbbm{1}_{\{i_h = i\}}$ \label{line:meank}
		\State $c_{t-1}^i \gets \overline{R}^i\sqrt{\frac{2\ln t}{n^i_{t-1}}}$ \label{line:ucbk}
		\State $u^i_{t-1} \gets \hat{R}^{i}_{t-1} + c_{t-1}^i $
		\EndFor
		\State Pull arm $i_t = \argmax_{i \in [K]}$ $u^i_{t-1}$ \label{line:pullucb}
		\State Observe the reward $r^{i_t}_t$ of the arm pulled at round $t$ \label{line:obsucb}
		\EndFor
	\end{algorithmic}
	\label{a:alpha_ucb1}
\end{algorithm}

\section{Baseline Algorithms Description} \label{S:baselines}

In this section, we report the details about the algorithms from the literature which we use as baselines in the experiments of Section~\ref{S:exp}. 
In particular, we compare the performances of the proposed \texttt{TP-UCB-FR} and the \texttt{TP-UCB-EW} with those of the baselines \texttt{UCB1}, assuming to obtain all the rewards corresponding to the pull of arm $i_t$ at time $t$, and \texttt{Delayed-UCB1}, which uses the realization of the pulls only when they are complete, i.e., with a constant delay of $\Tmax-1$.

\subsection{Non-Delayed Feedback} \label{s:UCB}

We describe the version of the \texttt{UCB1} algorithm, designed by~\citeauthor{auer2002finite}~[\citeyear{auer2002finite}], in which the reward $r^{i_t}_t$ provided by pulling arm $i_t$ at round $t$ is observed by the learner at time $t$.
We recall that this algorithm cannot be run in a TP-MAB setting, unless we are in the degenerate case $\Tmax = 1$. It rather represents a clairvoyant algorithm having the information of the rewards $r^i_t$ without any delay.
We denote its policy by $\mathfrak{U}_{{\texttt{UCB1}}}$.

The pseudo-code of the \texttt{UCB1} algorithm is reported in Algorithm~\ref{a:alpha_ucb1}. 
During the initialization phase, all the arms are pulled once (Line~\ref{line:initucb}).
Subsequently, at each round $t$, the learner computes the empirical mean of the cumulative rewards $\hat{R}^{i}_{t-1}$ collected up to round $t - 1$ (Line~\ref{line:meank}), where we denote by $n^i_{t-1} := \frac{1}{n_i}\sum_{h=1}^{t-1} \mathbbm{1}_{\{i_h = i\}}$ the number of times the arm $i$ has been pulled up to round $t - 1$, and the the confidence interval $c^i_{t-1}$ (Line~\ref{line:ucbk}).
Finally, the learner pulls the arm with the largest upper confidence bound $u^i_{t-1}$ (Line~\ref{line:pullucb}), and observes the reward $r^{i_t}_t$ (Line~\ref{line:obsucb}).

We provide the following upper bound on the regret of the \texttt{UCB1} algorithm (see the proof by~\cite{auer2002finite}):
\begin{restatable}{theorem}{NDalpha}\label{t:nd-alpha}
	The pseudo-regret of \texttt{UCB1} after $T \in \mathbb{N}^*$ rounds on a MAB problem with $r^i_t$ rewards is:
	\begin{equation*}
	    \mathcal{R}_T(\mathfrak{U}_{\emph{\texttt{UCB1}}}) \leq \sum_{i:\mu_i<\mu^*} \frac{8 ( \overline{R}^i)^2 \ln{T}}{\Delta_i} + \mleft(1 + \frac{\pi^2}{3}\mright)\sum_{i:\mu_i<\mu^*} \Delta_i.
    \end{equation*}
\end{restatable}

\subsection{Delayed Feedback} \label{s:delayed}

We show how to apply the \texttt{Delayed-UCB1} algorithm, provided by~\citeauthor{joulani2013online}~[\citeyear{joulani2013online}] and originally designed for the Delayed-MAB setting, to the TP-MAB setting.
In the TP-MAB problem, the realization of the cumulative reward $r^i_t$ is observed after $\Tmax - 1$ rounds from the pull of the arm.
As a consequence, one always waits for $\Tmax - 1$ rounds before collecting the reward from a pull.
This approach, corresponds to a delayed-feedback MAB setting in which the delay is known and deterministic.
After such a delay, the learner updates the policy $\mathfrak{U}_{{\texttt{D-UCB1}}}$ of \texttt{Delayed-UCB1} with the value of the cumulative reward.

The pseudo-code of the \texttt{Delayed-UCB1} algorithm applied to a generic TP-MAB setting is reported in Algorithm~\ref{a:ucb1_d}.
During the initialization phase, all arms are pulled in a round robin fashion until at least one reward is collected (Line~\ref{line:initducb}).
Subsequently, at each round $t$, the learner computes the empirical mean $\hat{R}^{i}_{t-1}$ of the cumulative rewards collected up to round $t-1$ (Line~\ref{line:meanducb}), where $s^i_{t-1} := \sum_{h=1}^{t-\Tmax} \mathbbm{1}_{\{i_h = i\}}$ is the number of complete reward observed so far for arm $i$, and the confidence interval $c^i_{t-1}$ (Line~\ref{line:ucbducb}).
Finally, the learner pulls the arm with the largest upper confidence bound $u^i_{t-1}$ (Line~\ref{line:pullducb}), and observes the reward corresponding to the pull occurred at round $t - \Tmax + 1$ (Line~\ref{line:obsducb}).
When no sample is available for an arm $i$ its upper bound is set to $+\infty$.
\begin{algorithm}[t!]
	\caption{\texttt{Delayed-UCB1}}	\label{a:ucb1_d}
	\begin{algorithmic}[1]
		\For { $t \in  \{1, \ldots, \Tmax\}$} \Comment{init phase}
		\State Pull arm $i_t = ((t-1)\mod{K})+1$
		\label{line:initducb}
		\EndFor
		\For { $t \in  \{\Tmax+1, \ldots, T\}$} \Comment{loop phase}
		\For { $i \in  \{1, \ldots, K\}$}
		\State $\hat{R}^i_{t-1} \gets \frac{1}{s^i_{t-1}}\sum_{h=1}^{t-\Tmax} r^i_{h} \mathbbm{1}_{\{i_h = i\}}$ \label{line:meanducb}
		\State $c^i_{t-1} \gets \overline{R}^i\sqrt{\frac{2 \ln (t-1)}{s^i_{t-1}}}$ \label{line:ucbducb}
		\State $u^i_{t-1} \gets \hat{R}^i_{t-1} + c^i_{t-1} $ 
		\EndFor
		\State Pull arm $i_t = \argmax_{i \in [K]}$ $u^i_{t-1}$ \label{line:pullducb}
		\State Observe reward $r^{i_{t-\Tmax+1}}_{t-\Tmax+1}$ of the arms pulled at round ${t-\Tmax+1}$ \label{line:obsducb}
		\EndFor	
	\end{algorithmic}
\end{algorithm}
We provide the following upper bound on the regret of the \texttt{Delayed-UCB1} algorithm (see~\citeauthor{joulani2013online}~[\citeyear{joulani2013online}]).
\begin{theorem}\label{t:d-alpha}
	The pseudo-regret of \texttt{Delayed-UCB1} after $T \in \mathbb{N}^*$ rounds in the TP-MAB setting is:
	\begin{equation}
	    \mathcal{R}_T(\mathfrak{U}_{\emph{\texttt{D-UCB1}}}) \le \sum_{i:\mu_i<\mu^*} \frac{8(\overline{R}^i)^2\ln{T}}{\Delta_i} + \mleft(1 + \frac{\pi^2}{3} + \Tmax \mright)\sum_{i:\mu_i < \mu^*} \Delta_i.
	\end{equation}
\end{theorem}
\begin{proof}
	The theorem follows from Theorem $7$ by~\cite{joulani2013online}, where the expected value of the maximum number of missing feedback of arm $i$ during the first $t$ time steps is $\mathbb{E}[G^*_{i,t}] < \Tmax$, where $G^*_{i,t}$ is the maximum number of missing feedbacks during the first $t$ rounds for arm $i$.
\end{proof}	

\subsection{Theoretical Results Summary}

Finally, we provide a table summarizing the results known in the literature and provided in this paper.
Table~\ref{tab:theoresults} reports the lower and upper bounds on the regret for different algorithms and settings. Notice that the lower bound results hold for $T \rightarrow +\infty$. Moreover, in Table~\ref{tab:theoresults}, we denote $\frac{8 ( \overline{R}^i)^2}{\Delta_i}$ by $C_i$ and $\sum_{i:\mu_i<\mu^*}$ by $\sum_{i}$.
The assumption is that the instantaneous (for the MAB and Delayed-MAB settings) and cumulative (for the TP-MAB setting) rewards have support in $[0, \overline{R}^i]$. Moreover, in the Delayed-MAB setting, the maximum stochastic delay is $\tau_{\max}$
The novel results have been highlighted in blue.
\texttt{UCB1} does not have guarantees in the Delayed-MAB and TP-MAB settings since it has been developed for a more restrictive scenario, i.e., $\tau_{\max} = 1$.

The results related to the proposed algorithms, i.e., \texttt{TP-UCB-FR} and \texttt{TP-UCB-EW}, for the MAB setting have been derived fixing $\tau_{\max} = 1$ and $\alpha = 1$ in the corresponding theorems. The results of the Delayed-MAB setting have been derived fixing $\alpha = 1$.
We remark that \texttt{TP-UCB-FR} in the MAB setting has the same asymptotic order of upper bound of \texttt{UCB1}, while the upper bound of \texttt{TP-UCB-EW} reduces exactly to the one of \texttt{UCB1} in this setting.

\begin{table}[t!]
\caption{Summary of the known theoretical results (in black) and original contributions provided in this paper (in blue).} \label{tab:theoresults}
\centering
\scriptsize
\begin{tabular}{|l||p{0.27\textwidth}|p{0.27\textwidth}|p{0.27\textwidth}|}
\hline
& \multicolumn{3}{c|}{Setting} \\
\hline
& MAB & Delayed-MAB  & TP-MAB with $\alpha$-smoothness \\
\Xhline{2\arrayrulewidth}
Lower bound & {\color{black} $\sum_{i} \frac{\Delta_i \ln T }{KL\left( \frac{\mu_i}{\overline{R}_{\max}}, \frac{\mu^*}{\overline{R}_{\max}} \right)}$} &  {\color{black} $\sum_{i} \frac{\Delta_i \ln T }{KL\left( \frac{\mu_i}{\overline{R}_{\max}}, \frac{\mu^*}{\overline{R}_{\max}} \right)}$} & {\color{blue} \boldmath $\sum_{i} \frac{\Delta_i \ln T}{ \alpha KL\left( \frac{\mu_i}{\overline{R}_{\max}}, \frac{\mu^*}{\overline{R}_{\max}} \right)}$} \\
\hline
\texttt{UCB1}       & $\sum_{i} C_i \ln{T} + \mleft(1 + \frac{\pi^2}{3}\mright)\sum_{i} \Delta_i$ & N.a. & N.a.\\
\hline
\texttt{Delayed-UCB1} & {\color{black} $\sum_{i} C_i \ln{T} + \mleft(1 + \frac{\pi^2}{3} \mright)\sum_{i} \Delta_i$} & {\color{black} $\sum_{i} C_i \ln{T} + \mleft(1 + \frac{\pi^2}{3} + \Tmax \mright)\sum_{i} \Delta_i$} & {\color{blue} \boldmath $\sum_{i} C_i \ln{T} + \mleft(1 + \frac{\pi^2}{3} + \Tmax \mright)\sum_{i} \Delta_i$}\\
\hline
\texttt{TP-UCB-FR} & {\color{blue} \boldmath $\sum_{i} \frac{C_i}{2} \ln T \mleft( 1 + \sqrt{1 + \frac{2\Delta_i}{\overline{R}^i \ln T}} \mright) +  2 \sum_{i} \overline{R}^i + \mleft( 1 + \frac{\pi^2}{3} \mright) \sum_{i} \Delta_i$} & {\color{blue} \boldmath $\sum_{i} \frac{C_i}{2}  \ln T \mleft( 1 + \sqrt{1 + \frac{2\Delta_i}{\overline{R}^i \ln T}} \mright) +  2 \tau_{\max} \sum_{i} \overline{R}^i + \mleft( 1 + \frac{\pi^2}{3} \mright) \sum_{i} \Delta_i$} & {\color{blue} \boldmath $\sum_{i} \frac{C_i}{2\alpha} \ln T \mleft( 1 + \sqrt{1 + \frac{\alpha(\alpha + 1) \Delta_i}{2 \overline{R}^i \ln T}} \mright) +  (\alpha+1) \step \sum_{i} \overline{R}^i + \mleft( 1 + \frac{\pi^2}{3} \mright) \sum_{i} \Delta_i$} \\
\hline
\texttt{TP-UCB-EW} & {\color{blue} \boldmath $\sum_{i} C_i  \ln T + \mleft(1 + \frac{\pi^2}{3}\mright) \sum_{i} \Delta_i$} & {\color{blue} \boldmath $ \sum_{i} C_i  \ln T + \mleft(\tau_{\max} +\frac{\pi^2}{3}\mright) \sum_{i} \Delta_i$} & {\color{blue} \boldmath $\sum_{i} C_i  \ln T + \alpha \mleft(\step +\frac{\pi^2}{3}\mright) \sum_{i} \Delta_i$}\\
\hline
\end{tabular}
\end{table}

\section{Omitted Proofs} \label{a:proof_UB}

\thmlower*

\begin{proof}

At first, notice that learning the optimal arm in a TP-MAB problem $\mathcal{P}$ for rewards $R^i_t$ taking values over a generic finite domain $[0, \overline{R}^i]$, having range $\overline{R}^i$, is equivalent to the problem of learning in a TP-MAB problem $\mathcal{P}'$ with reward $\frac{R^i_t}{\overline{R}_{\max}}$ having domain $[0, 1]$.
Indeed, from a learning perspective, distinguish between two arms in the first setting requires the same sample complexity of distinguish between two arms in the second one.
The expected reward of the $i$-th arm of the $\mathcal{P}'$ problem is $(\mu_i)' = \frac{\mu_i}{\overline{R}_{\max}}$ and the one corresponding to the optimal arm is $(\mu^*)' = \frac{\mu^*}{\overline{R}_{\max}}$.

Let us consider for each problem $\mathcal{P}$ in the class of TP-MAB problems, its corresponding $\mathcal{P}'$ one.
For each $\mathcal{P}'$, we build a corresponding Delayed-MAB equivalent problem, by delaying all the intermediate rewards corresponding to a pull at round $t$ to the round $t+\Tmax-1$.
Therefore, using the results on the lower bound of the Delayed-MAB problems provided by~\citeauthor{vernade2017stochastic}~[\citeyear{vernade2017stochastic}] (Lemma~$15$) we have that:
\begin{equation}
	\lim \inf_{T \rightarrow +\infty} \frac{\mathbb{E}[N_i(T)]}{\log(T)} \geq \frac{1}{KL\left( \frac{\mu_i}{\overline{R}_{\max}}, \frac{\mu^*}{\overline{R}_{\max}} \right)},
\end{equation}
where $\mathbb{E}[N_i(T)]$ is the expected number of times an arm $i$ is selected over a time horizon of $T$ by the policy $\mathfrak{U}$.
Due to the equivalence depicted above, this result holds also for the original problems $\mathcal{P}$ in the class of TP-MAB problems.
From the fact that $\mathcal{R}_T(\mathfrak{U}) = \Delta_i \mathbb{E}[N_i(T)]$ and summing over the suboptimal arms, i.e., $i \neq i^*$, we get the theorem statement.
\end{proof}

\thmloweralpha*
\begin{proof}
	
	The proof follows the steps provided for Theorem~$2.2$ in the work by~\citeauthor{bubeck2012regret}~[\citeyear{bubeck2012regret}] and generalize them to the setting in which multiple rewards, \emph{i.e.}, $\alpha$, are earned by a single arm pull.
	
	Let us define an auxiliary MAB setting in which:
	\begin{itemize}
		\item only two arms with expected value $\mu_1$ and $\mu_2$, with $\mu_2 < \mu_1 <1$;
		\item all the arm have maximum reward equal to $R^i_t = \overline{R}_{\max}$;
		\item the reward $Z^i_{t,k}$ are i.i.d. over $k \in \{1, \ldots, \alpha\}$, meaning that the expected value of each of the element is $\frac{\mu_i}{\alpha}$;
		\item the reward are $Z^i_{t,k} \in \{0, \frac{\overline{R}_{\max}}{\alpha} \}$, \emph{i.e.}, the reward are Bernoulli scaled by a factor $\frac{\overline{R}_{\max}}{\alpha}$;
		\item pulling an arm at time $t$ provides $\alpha$ reward for the arm $\{Z^i_{t,1}, \ldots, Z^i_{t,\alpha}\}$, all observed by the learner at the time of the pull.
	\end{itemize}
	Let us remark that determining the optimality of an arm in this problem is no harder than the one in which the reward is spread over the period $\{t, \ldots, t + \Tmax -1\}$.
	Therefore, the derivation of a lower bound for this problem would also provide a lower bound for the original TP-MAB setting with $\alpha$-smoothness.
	Moreover, let us recall that learning in a problem where the reward are scaled by a factor $\frac{\overline{R}_{\max}}{\alpha}$, similarly to what has been done in Theorem~\ref{thm:lower}, does not change the complexity of learning.
	From now on, we will consider as expected value of the two arms $\mu_{Z_1} := \frac{\mu_1}{\overline{R}_{\max}}$ and $\mu_{Z_2} := \frac{\mu_2}{\overline{R}_{\max}}$.
	Therefore, to compute the expected value of number of times an algorithm pulls the suboptimal arm $\mathbb{E}[N_2(T)]$ we can also use the scaled rewards.
	In what follows, we prove that the lower bound for the auxiliary problem for any uniformly efficient policy $\mathfrak{U}$.\ \\

	\textbf{Overall proof idea}
	
	Let us consider a second instance of the above defined MAB such that arm $2$ is optimal and $\mu_{Z_1} < \mu'_{Z_2} < 1$.
	We refer to it as the modified bandit.
	Let $\varepsilon > 0$, since $x \mapsto KL(\mu_{Z_2},x)$ is continuous one can find $\mu'_{Z_2} \in (\mu_{Z_1},1)$ such that:
	\begin{align}
		KL\mleft( \mu_{Z_2}, \mu'_{Z_2} \mright) \leq (1 + \varepsilon) KL(\mu_{Z_2},\mu_{Z_1}). \label{link}
	\end{align}
	In what follows, we use the notation $\mathbb{E}'$, $\mathbb{P}'$ to denote the expected value and probability computed in the second bandit instance. The goal is to compare the behavior of the forecaster on the initial and modified bandits. 
	The idea of the proof is to show that, with a big enough probability, the forecaster is not able to distinguish between the two problems.
	Then, using the fact that the forecaster is uniformly efficient by hypothesis, we show that the algorithm does not make too many mistake on the modified bandit and, in particular, provide a lower bound on the number of times the optimal arm is played.
	This reasoning implies a lower bound on the number of times the suboptimal arm $2$ is played in the initial problem.
	\ \\
	
	\textbf{First step: $\mathbb{P}(C_t) = o(1)$}\ \\
	Let us define, for $s \in \{1, \ldots, t\}$, the empirical estimate of $KL\mleft( \mu_{Z_2}, \mu'_{Z_2} \mright)$ at round $t$ when the arm $2$ is pulled $s$ times:
	\begin{align}
		\widehat{KL}_{\alpha s} := \sum_{n = 1}^{s} \sum_{k=1}^\alpha \ln \frac{\mu_{Z_2} Z^2_{n,k} + (1-\mu_{Z_2})(1-Z^2_{n,k})}{\mu'_{Z_2} Z^2_{n,k} + (1-\mu'_{Z_2})(1-Z^2_{n,k})}.
	\end{align}
	
	We introduce the following event linking the behavior of the forecaster on the initial and modified bandits:
	\begin{align}
		C_t := \mleft\{ \alpha N_2(t) < f_t \quad \textnormal{and} \quad  \widehat{KL}_{\alpha N_2(t)} \leq \mleft( 1 - \varepsilon/2 \mright) \ln t \mright\},
	\end{align}
	where $f_t = \frac{1- \varepsilon}{KL(\mu_{Z_2},\mu'_{Z_2})} \ln t$.
	Following the proof of Theorem~$2.2$ from~\citeauthor{bubeck2012regret}~[\citeyear{bubeck2012regret}], we have:
	\begin{equation} \label{eq:measurechange}
		\mathbb{P}'(C_t) = \mathbb{E}[ 1_{C_t} \exp{\mleft(- \widehat{KL}_{\alpha N_2(t)}\mright)  } ] \geq e^{- (1 - \varepsilon/2) \ln t} \mathbb{P}(C_t),
	\end{equation}
	where we used the change of measure identity for the first equality and use the fact that $\widehat{KL}_{\alpha N_2(t)} \leq \mleft( 1 - \varepsilon/2 \mright) \ln t$ in $C_t$.\footnote{For any event $A$ in the $\sigma$-algebra generated by $\{Z^2_{n,k} \}_{n \in \{1, \ldots, s\}, k \in \{1, \ldots, \alpha\}}$ holds that $\mathbb{P}'(A) =  \mathbb{E} \mleft[ 1_{A} \exp{\mleft(- \widehat{KL}_{\alpha N_2(t)}\mright)} \mright]$.}
	Combining Equation~\eqref{eq:measurechange}, the definition of $C_t$, and using the Markov's inequality, we have:
	\begin{align}
		\mathbb{P}(C_t) \leq t^{(1-\varepsilon/2)} \mathbb{P}'(C_t) \leq t^{(1-\varepsilon/2)} \mathbb{P}'(\alpha N_2(t) < f_t) \leq t^{(1-\varepsilon/2)} \frac{\mathbb{E}' \left[t-N_2(t)\right]}{t-f_t/\alpha} = o(1),
	\end{align}
	where with $o(1)$ we denote a quantity whose limit for $t \rightarrow +\infty$ is $0$ and we used the fact that the policy $\mathfrak{U}$ is uniformly efficient, \emph{i.e.}, $\mathbb{E}'[T_2(t)] = o(t^\beta)$ with $\beta < 1$.
	\ \\
	
	\textbf{Second step: $\mathbb{P}(\alpha N_2(t) \leq f_t) = o(1)$}\ \\
	Using the Third step of the proof of Theorem~$2.2$ from~\citeauthor{bubeck2012regret}~[\citeyear{bubeck2012regret}], we get:
	\begin{equation}
		o(1) = \mathbb{P}(C_t) \leq \mathbb{P} \left( \underbrace{\alpha T_2(t) < f_t}_{E_1} \wedge \underbrace{\frac{KL(\mu_{Z_2}, \mu'_{Z_2})}{(1 - \varepsilon) \ln t} \max_{s < f_t/\alpha} \widehat{KL}_{\alpha s} \leq \frac{1 - \varepsilon/2}{1 - \varepsilon} KL(\mu_{Z_2}, \mu'_{Z_2})}_{E_2} \right).
	\end{equation}
	Using the strong law of large numbers the event $E_2$ is s.t.~$\lim_{t \rightarrow +\infty} \mathbb{P}(E_2) =1$, we infer that $\mathbb{P}(E_1) = \mathbb{P}(\alpha N_2(t) < f_t) = o(1)$, and that for $t \rightarrow + \infty$ we have $\mathbb{E}[N_2(t)] > f_t/\alpha$.
	\ \\
	
	\textbf{Final step}\ \\
	Using Equation~\eqref{link} we have that, for $t \rightarrow + \infty$: 
	\begin{align}
		\mathbb{E}[N_2(t)] > f_t / \alpha = \frac{1- \varepsilon}{\alpha KL(\mu_{Z_2},\mu'_{Z_2})} \ln t \geq \frac{1- \varepsilon}{\alpha (1 + \varepsilon) KL(\mu_{Z_2},\mu_{Z_1})} \ln t,
	\end{align}
	where the theorem statement follows from the arbitrarity of the value of $\varepsilon$, substituting $\mu_{Z_1}$ with $\frac{\mu^*}{\overline{R}_{\max}}$ and $\mu_{Z_2}$ with $\frac{\mu_2}{\overline{R}_{\max}}$, and summing over all the suboptimal arms.
	
\end{proof}
	
\FRreg*

\begin{proof}
Let us define the true empirical mean of the cumulative reward of arm $i$ computed over $n^i_{t}$ samples as follows:
\begin{equation*}
    \hat{R}^{i,\textnormal{true}}_{t} := \frac{1}{n^i_{t}} \sum_{h=1}^{t} r^i_h \mathbbm{1}_{\{i_h = i\}}.
\end{equation*}
We aim to bound the difference between $\hat{R}^{i,\textnormal{true}}_{t}$ and the approximated empirical mean of the cumulative reward $\hat{R}^i_{t}$ from arm $i$ computed over $n^i_{t}$ samples as in the \texttt{TP-UCB-FR} algorithm.
Formally, we have:
\begin{align}
    \hat{R}^{i,\textnormal{true}}_{t} - \hat{R}^{i}_{t} &= \frac{1}{n^i_{t}} \sum_{h=1}^{t} \sum_{j=1}^{\Tmax} \mleft(x^i_{h,j} - \tilde{x}^i_{h,j}\mright) \mathbbm{1}_{\{i_h = i\}} \leq \frac{1}{n^i_{t}} \sum_{h=1}^{t} \sum_{j=1}^{\Tmax} \mleft(x^i_{h,j} - \tilde{x}^i_{h,j}\mright) \\
    & = \frac{1}{n^i_{t}} \sum_{h=\max\{1,t-\Tmax+2\}}^{t} \sum_{j=t-h+2}^{\Tmax} x^i_{h,j} \label{eq:pos}\\
    & \leq \frac{1}{n^i_{t}} \sum_{j=1}^\alpha \step j \frac{\overline{R}^i}{\alpha} \label{eq:bousmo}\\
    & = \frac{\phi}{n^i_{t}} \frac{\overline{R}^i}{\alpha} \sum_{j=1}^{\alpha} j = \frac{\phi}{n^i_{t}} \frac{\overline{R}^i}{\alpha} \frac{\alpha(\alpha+1)}{2} =
    \frac{\overline{R}^i(\alpha+1) \step}{2 n^i_{t}},
\end{align}
where, Equation~\eqref{eq:pos} is due to the fact that $\underline{R}^i = 0$ for each $i \in [K]$, and the inequality in Equation~\eqref{eq:bousmo} is due to the $\alpha$-smoothness of the environment.

Following the proof of Theorem $1$ by~\citeauthor{auer2002finite}~[\citeyear{auer2002finite}], we bound the expected number of time a suboptimal arm is pulled as follows:
\begin{equation} \label{eq:subopt}
    \mathbb{E}[N_i(t)] \le \ell + \sum_{t=1}^{\infty}\sum_{s=1}^{t-1}\sum_{s_i=\ell}^{t-1} \mathbb{P} \mleft\{\mleft(\hat{R}^{*}_{t,s} + c_{t,s}^*\mright) \le \mleft(\hat{R}^{i}_{t,s_i} + c_{t,s_i}^i\mright) \mright\},
\end{equation}
where $\hat{R}^{*}_{t,s}$ and $c_{t,s}^*$ are the empirical mean computed as in the \texttt{TP-UCB-FR} algorithm and the confidence bound, respectively, of the optimal arm $i^*$ in the case $s$ pulls occurred in the first $t$ rounds, and, $\hat{R}^{i}_{t,s_i}$ and $c_{t,s_i}^i$ are the empirical mean computed as in the \texttt{TP-UCB-FR} algorithm and the confidence bound, respectively, of the arm $i$ in the case $s_i$ pulls occurred in the first $t$ rounds.

Equation~\eqref{eq:subopt} implies that at least one of the following holds:
\begin{align}
    &\hat{R}^{*}_{t,s} \leq \mu^* - c_{t,s}^*, \label{eq:optisub}\\
    &\hat{R}^{i}_{t,s_i} \geq \mu_i + c_{t,s_i}^i, \label{eq:suboptover}\\
    &\mu^* < \mu_i + 2c_{t,s_i}^i. \label{eq:ineq_false}
\end{align}

Let us focus on Equation~\eqref{eq:optisub}.
We have that:
\begin{align}
	&\mathbb{P} \mleft(\hat{R}^{*}_{t,s} - \mu^{*} \leq - c_{t,s}^* \mright)  = \mathbb{P}\mleft(\hat{R}^{*,\textnormal{true}}_{t,s} - \mu^* \leq - c_{t,s}^* + \hat{R}^{*,\textnormal{true}}_{t,s} - \hat{R}^{*}_{t,s} \mright) \\
	& \leq \mathbb{P}\mleft(\hat{R}^{*,\textnormal{true}}_{t,s} - \mu^* \leq - c_{t,s}^* + \frac{\overline{R}^i(\alpha+1) \step}{2 s} \mright) = \mathbb{P}\mleft(\hat{R}^{*,\textnormal{true}}_{t,s} - \mu^* \leq -\overline{R}^* \sqrt{\frac{2\ln t}{\alpha s}} \mright) \\
	& \leq \exp \left\{ \frac{ 2 \mleft( \overline{R}^* \sqrt{\frac{2\ln t}{\alpha s}} \mright)^2 s^2}{\sum_{l = 1}^{\alpha s} \mleft(\frac{\overline{R}^*}{\alpha}\mright)^2} \right\} \leq e^{- 4 \ln t} \leq t^{-4}, \label{eq:hoe1}
\end{align}
where $c^*_{t,s} := \overline{R}^* \sqrt{\frac{2 \ln (t)}{\alpha s}} + \frac{ \overline{R}^i(\alpha+1)\step}{2 s}$, $\overline{R}^* := \overline{R}^{i^*}$, $\hat{R}^{*,\textnormal{true}}_{t,s}$ is the empirical mean of the optimal arm $i^*$ in the case $s$ pulls occurred in the first $t$ rounds, and we use the Hoeffding inequality in Equation~\eqref{eq:hoe1}.

Similarly, Equation~\eqref{eq:suboptover} is bounded by:
\begin{align}
	&\mathbb{P} \mleft(\hat{R}^{i}_{t,s_i} - \mu_{i} \geq c_{t,s_i}^i \mright) \leq \mathbb{P}\mleft(\hat{R}^{i,\textnormal{true}}_{t,s} - \mu_{i} \geq \overline{R}^i \sqrt{\frac{2\ln t}{\alpha s_i}} \mright) \\
	& \leq e^{-4\ln t}=t^{-4}, \label{eq:hoe2}
\end{align}
where we used the fact that $\hat{R}^{i,\textnormal{true}}_{t,s_i} \geq \hat{R}^{i}_{t,s}$ by construction of the latter, and we used the Hoeffding inequality to derive Equation~\eqref{eq:hoe2}.

Define:
\begin{equation} \label{eq:ell}
    \ell := \mleft \lceil \frac{\overline{R}^i(\alpha+1)\phi}{\Delta_i} + \frac{4(\overline{R}^i)^2 \ln t}{\alpha \Delta_i^2} \mleft(1 + \sqrt{1+\frac{\alpha(\alpha+1)\phi\Delta_i}{2\overline{R}^i \ln t}}\mright) \mright\rceil.
\end{equation}
We have that the following holds:
\begin{align*}
    &\mu^* \geq \mu_i + 2c_{t,s}^i \\
    &\Delta_i \geq 2 \mleft(\overline{R}^i\sqrt{\frac{2 \ln t}{\alpha s_i}} + \step \frac{\overline{R}^i (\alpha+1)}{2 s_i} \mright) \\
    &s_i^2 \mleft(\frac{\Delta_i^2}{4}\mright) - 2s_i \mleft(\frac{\Delta_i\overline{R}^i(\alpha+1)}{4} \step + \frac{(\overline{R}^i)^2 \ln t}{\alpha}\mright) + \step^2 \frac{(\overline{R}^i)^2 (\alpha+1)^2}{4} \geq 0 \\
    & s_i \geq \frac{4}{\Delta_i^2} \mleft(\frac{\Delta_i\overline{R}^i(\alpha+1)}{4}\step + \frac{(\overline{R}^i)^2\ln t}{\alpha} + \sqrt{ \frac{(\overline{R}^i)^4 \ln^2 t}{\alpha^2} + \frac{\Delta_i (\overline{R}^i)^3 (\alpha+1)\phi \ln t}{2 \alpha}} \mright)\\
    & s_i \geq \frac{\overline{R}^i(\alpha+1)}{\Delta_i}\step + \frac{4(\overline{R}^i)^2 \ln t}{\Delta_i^2\alpha} \mleft(1 + \sqrt{1+ \frac{\Delta_i \alpha(\alpha+1)\phi}{2\overline{R}^i \ln t}}\mright),
\end{align*}
and, therefore, for $s_i \ge \ell$ the inequality in Equation~\eqref{eq:ineq_false} is always false.

Finally, summing up the results derived above and using $\ell$ as defined in Equation~\eqref{eq:ell}, we have:
\begin{align}
    & \mathbb{E}[N_i(t)] \leq \mleft\lceil \frac{\overline{R}^i(\alpha+1)}{\Delta_i} \step + \frac{4(\overline{R}^i)^2 \ln t}{\alpha \Delta_i^2} \mleft(1 + \sqrt{1+\frac{\alpha(\alpha+1)\phi\Delta_i}{2\overline{R}^i \ln t}}\mright) \mright\rceil \\
    & + \sum_{t=1}^{\infty}\sum_{s=1}^{t-1}\sum_{s_i=\ell}^{t-1} \left[ \mathbb{P} \left( \hat{R}^{*}_{t,s} - \mu^* \leq - c_{t,s}^* \right) + \mathbb{P} \left( \hat{R}^{i}_{t,s_i} -\mu_i \geq c_{t,s_i}^i \right) \right]\\
    & \leq 1 + \frac{\overline{R}^i(\alpha+1)}{\Delta_i} \step + \frac{4(\overline{R}^i)^2 \ln t}{\alpha \Delta_i^2} \mleft(1 + \sqrt{1+\frac{\alpha(\alpha+1)\phi\Delta_i}{2\overline{R}^i \ln t}}\mright) + 1 + \sum_{t=1}^{\infty}\sum_{s=1}^{t-1}\sum_{s_i=\ell}^{t-1} 2 t^{-4} \\
    & \leq \frac{\overline{R}^i(\alpha+1)}{\Delta_i}\step + \frac{4(\overline{R}^i)^2 \ln t}{\alpha \Delta_i^2} \mleft(1 + \sqrt{1+\frac{\alpha(\alpha+1)\phi\Delta_i}{2\overline{R}^i \ln t}}\mright) + 1 + \frac{\pi^2}{3}.
\end{align}

The theorem statement follows by the fact that $\mathcal{R}_T(\mathfrak{U}_{{\texttt{FR}}}) = \sum_{i : \mu_i < \mu^*} \Delta_i \mathbb{E}[N_i(T)]$.
\end{proof}

\EWreg*

\begin{proof}
Following the same proof strategy of Theorem~\ref{t:FRreg}, we want to bound the expected value of the number of pulls of suboptimal arms:
\begin{equation} \label{eq:probtot}
    \mathbb{E}[N_i(t)] \leq l + \sum_{t=1}^{\infty}\sum_{s=1}^{t-1} \sum_{s_i=l}^{t-1} \mathbb{P} \mleft( \sum_{k=1}^{\alpha}(\hat{Z}^*_{t,k,s} + c^*_{t,k,s}) \leq \sum_{k=1}^{\alpha}(\hat{Z}^i_{t,k,s_i} + c^i_{t,k,s_i}) \mright),
\end{equation}	
where $\hat{Z}^*_{t,k,s}$ and $c^*_{t,k,s}$ are the empirical mean computed as in the \texttt{TP-UCB-EW} algorithm and the confidence bound, respectively, of the optimal arm $i^*$ in the case $s$ pulls occurred in the first $t$ rounds, and, $\hat{Z}^i_{t,k,s_i}$ and $c^i_{t,k,s_i}$ are the empirical mean computed as in the \texttt{TP-UCB-EW} algorithm and the confidence bound, respectively, of the arm $i$ in the case $s_i$ pulls occurred in the first $t$ rounds.
Notice that in this case the number of samples collected from each one of the $\alpha$ aggregated rewards are $\leq s$ and $\leq s_i$, respectively.
Moreover, for values of $l > \Tmax$ the quantities regarding the suboptimal arm are estimated using at least one sample, \emph{e.g.}, $0 < n^i_{t,k,s} < s_i$.
Conversely, for $s \leq \Tmax$ the optimal arm might have no sample available to estimate the expected value and the bound.
However, since the values of the upper confidence bound is set $+\infty$ if no sample is collected, the probability that it is smaller than the one of a suboptimal arm is $0$, (\emph{i.e.}, $\mathbb{P} \mleft( \sum_{k=1}^{\alpha}(\hat{Z}^*_{t,k,s} + c^*_{t,k,s}) \leq \sum_{k=1}^{\alpha}(\hat{Z}^i_{t,k,s_i} + c^i_{t,k,s_i}) \mright) = 0$). As a consequence, the cases in which no sample is available for the optimal bound can be disregarded.

The condition above is satisfied if at least one of the following $2\alpha+1$ inequalities holds:
\begin{align}
	&\hat{Z}^*_{t,k,s}-\mu^*_k \leq -c^*_{t,k,s}, & \forall k \in \{1, \ldots, \alpha\} \label{eq:opt2}\\
	&\hat{Z}^i_{t,k,s_i}-\mu_{i,k} \ge c^i_{t,k,s_i}, &\forall k \in \{1, \ldots, \alpha\} \label{eq:arm2} \\
	&\sum_{k=1}^{\alpha} \mu_{k}^* - \mu_{i,k} - 2c^i_{t,k,s_i} < 0 , \label{c:false}
\end{align}
where $\mu_{i,k} := \mathbb{E}[Z^i_{t,k,s_i}]$ and $\mu^*_k : = \mathbb{E}[Z^{*}_{t,k,s}]$ are the expected value of the aggregated reward $Z^i_{t,k,s_i}$ from arm $i$, and $Z^{i^*}_{t,k,s}$ from the optimal arm, respectively.

Let us focus on the $k$-th inequality in Equation~\eqref{eq:opt2}.	
We have:
\begin{align}
	&\mathbb{P}(\hat{Z}^*_{t,k,s} - \mu^*_k \leq -c^*_{t,k,s}) \leq \exp \left\{ -\frac{2 (n^*_{t,k,s})^2 (c^*_{t,k,s})^2}{\sum_{l=1}^{n^*_{t,k,s}} \left( \frac{\overline{R}^*}{\alpha} \right)^2} \right\} \\
	&\leq \exp \mleft\{ -\frac{2n^*_{t,k,s} (c^*_{t,k,s})^2 \alpha^2}{(\overline{R}^*)^2} \mright\} \leq e^{-4\ln{t}} \le t^{-4}, \nonumber
\end{align}
where $n^*_{t,k,s}$ is the number of samples available for the estimation of the expected value of $Z^{*}_{t,k,s}$ if we pulled $s$ times the arm $i^*$ at round $t$.
Here, we assume that the estimates have at least one sample.
If no samples are available, the original probability in Equation~\eqref{eq:probtot} is bounded by $0$.

Similarly, for the inequalities in Equation~\eqref{eq:arm2}, we have:
\begin{align}
	&\mathbb{P}(\hat{Z}^i_{t,k,s_i} - \mu_{i,k} \ge c^i_{t,k,s_i}) \leq \exp\mleft\{ -\frac{2(n^i_{t,k,s_i})^2(c^i_{t,k,s_i})^2}{\sum_{l=1}^{n^i_{t,k,s_i}}(\frac{\overline{R}^i}{\alpha})^2}\mright\}\\
	&\leq \exp \mleft\{ -\frac{2n^i_{t,k,s_i} (c^i_{t,k,s_i})^2 \alpha^2}{(\overline{R}^i)^2} \mright\} \leq e^{-4\ln{t}} \le t^{-4}.
\end{align}
where $n^i_{t,k,s_i}$ is the number of samples available for the estimation of the expected value of $Z^{i}_{t,k,s_i}$ if we pulled $s_i$ times the arm $i$ at round $t$.

Define $ l = \mleft \lceil \alpha\step - 1 + \frac{8(\overline{R}^i)^2\ln{t}}{\Delta_i^2} \mright \rceil$.
Notice that $l \geq \Tmax$.
We have that the inequality in Equation~\eqref{c:false} is false.
Indeed, we have that:
\begin{align}
	& \sum_{k=1}^{\alpha} \mleft(\mu_{k}^* - \mu_{i,k} - 2 \frac{\overline{R}^i}{\alpha}\sqrt{\frac{2\ln t}{n^i_{t,k,s_i}}} \mright)
	\geq \Delta_i - 2\frac{\overline{R}^i}{\alpha} \sum_{k=1}^{\alpha} \sqrt{\frac{2\ln t}{s_i - k \step + 1}} \nonumber \\
	&  \geq \Delta_i - 2\alpha\frac{\overline{R}^i}{\alpha} \sqrt{\frac{2\ln t}{s_i - \alpha \step + 1}} =  \Delta_i - 2\overline{R}^i \sqrt{\frac{2\ln{t}}{s_i - \alpha \step + 1}}, \label{c:zero}
	\end{align}
where we used that $\sum_{k=1}^{\alpha} \mu_{k}^* - \mu_{i,k} = \mu^*-\mu_i= \Delta_i$.

If $s_i \geq \alpha \step - 1 + \frac{8(\overline{R}^i)^2\ln t}{\Delta_i^2}$, we have that:
\begin{align}
	&s_i \geq \alpha \step - 1 + \frac{8 (\overline{R}^i)^2 \ln(t)}{\Delta^2_i}\\
	& \frac{\Delta_i^2}{4(\overline{R}^i)^2} \geq \frac{2\ln(t)}{s_i - \alpha \step + 1} \\
	&\Delta_i - 2\overline{R}^i \sqrt{\frac{2\ln(t)}{s_i - \alpha \step + 1}} \ge 0,
\end{align}
which implies that the inequality in Equation~\eqref{c:zero} is false.

Finally, summing the above results we have that:
\begin{align}
	& \mathbb{E}[N_i(t)] \leq \mleft \lceil \alpha \step - 1 + \frac{8(\overline{R}^i)^2 \ln(t)}{\Delta_i^2} \mright \rceil \nonumber\\
	& + \sum_{t=1}^{\infty}\sum_{s=1}^{t-1}\sum_{s_i=l}^{t-1}\sum_{k=1}^{\alpha}
	\left[ \mathbb{P}(\hat{Z}^*_{t,k,s} - \mu^*_k \leq - c^*_{t,k,s}) +
	\mathbb{P}(\hat{Z}^i_{t,k,s_i} - \mu_{i,k} \geq c^i_{t,k,s_i}) \right]\\
	& \le \alpha \step + \frac{8(\overline{R}^i)^2\ln(t)}{\Delta_i^2} + \sum_{t=1}^{\infty}\sum_{s=1}^{t-1}\sum_{s_i=l}^{t-1} 2\alpha t^{-4} \nonumber \\
	& \le \frac{8(\overline{R}^i)^2\ln t}{\Delta_i^2} + \alpha \mleft(\step +\frac{\pi^2}{3}\mright).
	\end{align}
Recalling that $\mathcal{R}_T(\mathfrak{U}_{{\texttt{EW}}}) = \sum_{i : \mu_i < \mu^*} \Delta_i \mathbb{E}[N_i(T)]$, concludes the proof.
\end{proof}

\clearpage

\section{Experimental Settings Description and Additional Experiments}\label{a:experiments}

\subsection{Technical Details}
The code has been run on a server equipped with Intel(R) Xeon(R) CPU $E5-4610$ v2 @ $2.30GHz$ and $126$ GiB of memory.
The operating system was Ubuntu $16.04.3$ LTS, and the experiments have been run on Python $3.5.2$.
The libraries used in the experiments, with the corresponding version were:
\begin{itemize}
    \item \texttt{numpy == 1.11.3}
    \item \texttt{tqdm == 4.14.0}
    \item \texttt{scipy == 0.18.1}
    \item \texttt{pandas == 0.20.3}
    \item \texttt{matplotlib == 3.3.4}
    \item \texttt{tikzplotlib == 0.9.8}
\end{itemize}

For the experiments the total time spent was $\approx 468$ hours, where the generation of the parameters of the synthetic datasest took $\approx 27$ hours, the execution of the algorithms for Setting \#1 $\approx 50$ hours, the execution of the algorithms for Setting \#2 $\approx 320$ hours, the execution of the algorithms for Spotify Setting $\approx 5$ hours (considering the data preprocessing operations), the execution of the algorithms for Setting \#4, presented in what follows, $\approx 66$ hours.

\subsection{Experimental Settings} \label{s: env}

In what follows, we provide a detailed description of those setting which have been presented in Section~\ref{S:exp} and further experiments confirming what has been showed in the main paper.

\begin{figure}[t!]
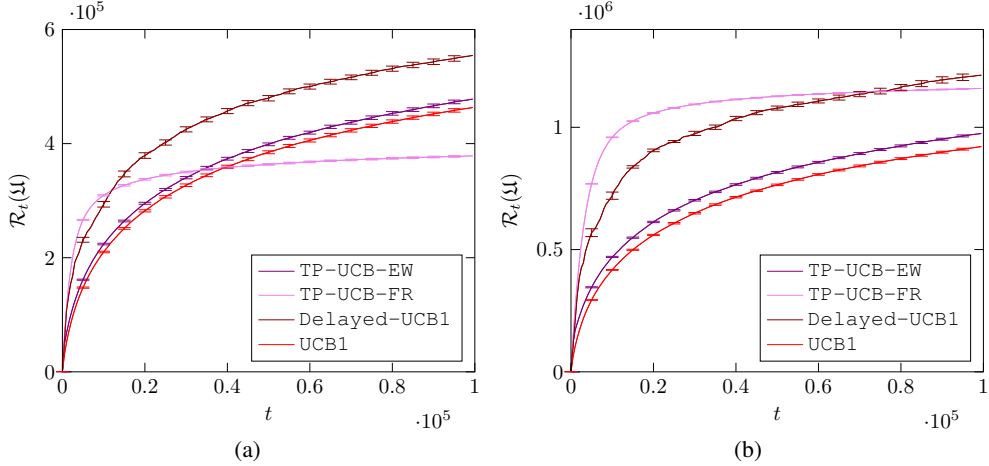

    \centering
    \subfloat[]{\input{images/experiment_2_100_10_uniform_mod}}
    \subfloat[]{\input{images/experiment_2_200_20_uniform_mod}}
    \caption{Experiments for Setting \#2 and uniform reward distribution: (a) $\Tmax = 100$, $\alpha = 10$, (b) $\Tmax = 200$, $\alpha = 20$.} \label{fig:unif}
\end{figure}

\paragraph{Setting \#2 (main paper scenario)}
In this setting, each arm is described  by a maximum reward $\overline{R}^i$ and two vectors $\bm{a}^i := \mleft[a^i_{1},\ldots,a^i_{\alpha}\mright]$ and $\bm{b}^i := \mleft[b^i_{1},\ldots,b^i_{\alpha}\mright]$ of length $\alpha$.
The aggregated reward $Z^i_{t,k}$ are distributed as $\mathcal{D}_k^i = \frac{\overline{R}^i}{\alpha}Beta(a_k^i,b_k^i)$, $\forall k \in [\alpha]$.
The results presented in the main paper are those corresponding to $\bm{a}^i := \bm{1}_{\alpha}$ and $\bm{b}^i := \bm{1}_{\alpha}$, where $\bm{1}_{\alpha}$ is a vector of length $\alpha$ whose elements are all $1$.
This setting corresponds to a uniform distribution over $\frac{\overline{R}^i}{\alpha}$ for each variable $Z^i_{t,k}$.
The corresponding results are presented in Section~\ref{S:exp}.
The regret over the entire time horizon is presented in Figure~\ref{fig:unif}.

\paragraph{Spotify Setting}

The original Spotify dataset~\cite{brost2019music} consists of listening sessions with levels of appreciation for each song associated to a user on the Spotify service.
Each listening session is truncated to $20$ tracks (songs).
Each row corresponds to the playback of one track pertaining to a specific listening session.
The dataset describes how users sequentially interact with the streamed content they are presented with.
More precisely, it contains information about when a user skips the playback of a track.

We preprocessed the available data as follows.
At first, for computational reasons we analysed only a fraction of the Spotify dataset.
Since we are interested in the listening sessions linked to a playlist, from that initial dataset we drop all the data associated with a \texttt{context\_type} field context which is different from \emph{editorial playlist}.
Moreover, we discarded all the listening sessions with less than $20$ songs and/or the user changed playlist during a single listening session (\texttt{context\_switch = true}).
This way, each listening sessions is composed of $20$ song coming from a single playlist.
We selected the $6$ most listened playlists having no overlapping songs, and extracted from the dataset the listening sessions corresponding to them.
The final dataset is available in the file \texttt{Spotify/spotifydf\_012.csv} in the code provided in the supplementary material.

The process of recommending the playlists is modeled as follows.
\begin{example}[Playlist Recommendation Problem - Reprise]

When a new user accesses the system, a playlist is proposed. 
This action corresponds to the selection of an arm $i$ by the recommendation algorithm.
The user will start the reproduction of the playlist, composed of exactly $N = 20$ songs.
For each song, at any time, the agent could decide to skip to the next song until the end of the playlist.
We aim at finding the playlist that maximizes the overall listening time.
Each song has a reward equal to \texttt{skip\_1}, \texttt{skip\_2}, \texttt{skip\_3}, and \texttt{not\_skipped}, representing increasing level of interest from the user.
These levels corresponds to the the realization of instantaneous reward $X^i_{t,j}$ of Bernoulli r.v.~that takes the value of $1$ if the user has reached at least the corresponding level and $0$ otherwise; 
The vector $\bm{X}^i_{t}$ has size equal to the number of songs of a playlist (\emph{i.e.}, aggregated rewards) times the number instant rewards returned by a song (\emph{i.e.}, $\phi$), and in this case $\Tmax = 20 \times 4 = 80$.
A summary of the expected rewards of the different playlists is provided in Table~\ref{tabSS}.
Figure~\ref{bucket_spotify} shows an example of the reproduction of part of $5$ songs of a playlist.
Songs $1$ and $3$ were listened completely, while Song $2$ was listened up to level the \texttt{skip\_2}.
Song $4$ and Song $5$ were entirely skipped.
\end{example}

\begin{figure}[t!]	
\centering
\begin{tabular}{cccccccccccccccccccc}
	\hline
	\multicolumn{1}{|c}{1} & 1 & 1 & \multicolumn{1}{c|}{1} & 1 & 1 & 0 & \multicolumn{1}{c|}{0} & 1 & 1 & 1 & \multicolumn{1}{c|}{1} & 0 & 0 & 0 & \multicolumn{1}{c|}{0} & 0 & 0 & 0 & \multicolumn{1}{c|}{0} \\ \hline
	\multicolumn{4}{c}{Song 1}                              & \multicolumn{4}{c}{Song 2}         & \multicolumn{4}{c}{Song 3}         & \multicolumn{4}{c}{Song 4}         & \multicolumn{4}{c}{Song 5}        
\end{tabular}
\caption{Example of a realization of an a subset of a playlist in the Spotify Setting.}
\label{bucket_spotify}
\end{figure}

\begin{table}[t!]
\centering
\caption{Description of the arms in the Spotify Setting.}
\label{tabSS}	
	\begin{tabular}{c|cccccc|}
		& $i=1$ & $i=2$ & $i=3$ & $i=4$ & $i=5$ & $i=6$  \\
		\hline 
		$\mu^i$    & 38.59 & 52.35 &  38.44 & 43.89 & 23.48& 36.20  \\ 
		$\sigma^i$ & 21.83 & 20.11  &23.09  & 23.14 & 23.48 & 23.8    \\ \hline
	\end{tabular}
\end{table}

\subsection{Additional Experiments}

\paragraph{Setting \#2.1}
In this experiment, the setting is the same as the one in Setting \#2, except that we designed the rewards s.t.~the first aggregated rewards after the pull are smaller than the last ones.
Specifically, the distribution are defined by the following vectors:
\begin{itemize}
    \item $\Tmax = 100$, $\alpha = 10$:
    \begin{align*}
        &\bm{a}^i=[2,4,6,8,10,10,10,10,10,10]; \\ &\bm{b}^i=[10,10,10,10,10,10,8,6,4,2];
    \end{align*}
    \item $\Tmax = 200$, $\alpha = 20$:
    \begin{align*}
    &\bm{a}^i=[2,4, \ldots, 18, 20, \ldots, 20]; \\ 
    &\bm{b}^i=[20, \ldots, 20, 18, \ldots, 4, 2];
    \end{align*}
    \item $\Tmax = 100$, $\alpha = 50$:
    \begin{align*}
    \bm{a}^i=[&2,4, \ldots, 48,50, \ldots, 50]; \\ 
    \bm{b}^i=[&50, \ldots, 50,48, \ldots, 4, 2];
    \end{align*}
    \item $\Tmax = 200$, $\alpha = 100$:
    \begin{align*}
        \bm{a}^i=[&2, 4, \ldots, 98, 100, \ldots, 100]; \\ 
        \bm{b}^i=[&100, \ldots, 100, 98, \ldots, 4, 2].
    \end{align*}
\end{itemize}

\begin{figure}[t!]
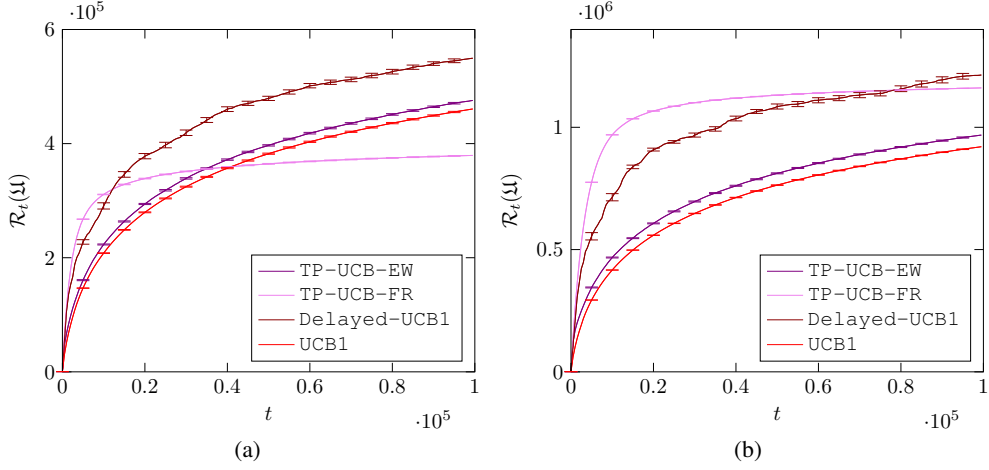

    \centering
    \subfloat[]{\input{images/experiment_2_100_10_1_mod}}
    \subfloat[]{\input{images/experiment_2_200_20_1_mod}}
    \caption{Experiments for Setting \#2.1: (a) $\Tmax = 100$, $\alpha = 10$, (b) $\Tmax = 200$, $\alpha = 20$.} \label{fig:early}
\end{figure}

The corresponding results are provided in Figure~\ref{fig:early}.
They are in line with the ones of Setting \#2.

\paragraph{Setting \#2.2}
In this experiment, the setting is the same as the one in Setting \#2, except that we designed the rewards s.t.~the first aggregated rewards after the pull are larger than the last ones.

Specifically, the distribution are defined by the following vectors:
\begin{itemize}
    \item $\Tmax = 100$, $\alpha = 10$:
    \begin{align*}
        &\bm{a}^i=[10,10,10,10,10,10,8,6,4,2,]; \\ &\bm{b}^i=[2,4,6,8,10,10,10,10,10,10];
    \end{align*}
    \item $\Tmax = 200$, $\alpha = 20$:
    \begin{align*}
        &\bm{a}^i=[20, \ldots, 20, 18, \ldots, 4, 2];\\
        &\bm{b}^i=[2,4, \ldots, 18, 20, \ldots, 20]; 
    \end{align*}
    \item $\Tmax = 100$, $\alpha = 50$:
    \begin{align*}
        \bm{a}^i=[&50, \ldots, 50,48, \ldots, 4, 2];\\
        \bm{b}^i=[&2, 4, \ldots, 48,50, \ldots, 50];
    \end{align*}
    \item $\Tmax = 200$, $\alpha = 100$:
    \begin{align*}
        \bm{b}^i=[&100, \ldots, 100, 98, \ldots, 4, 2];\\
        \bm{a}^i=[&2, 4, \ldots, 98, 100, \ldots, 100]. 
    \end{align*}
\end{itemize}

\begin{figure}[t!]
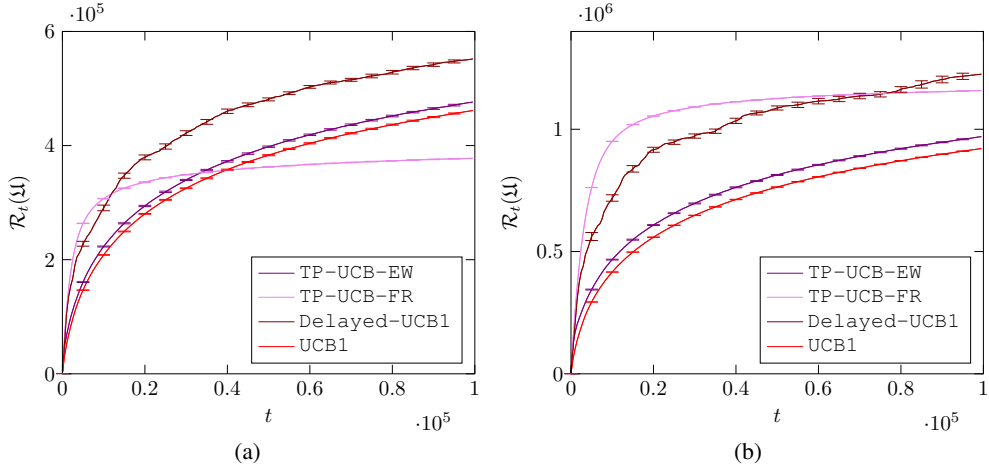

    \centering
    \subfloat[]{\input{images/experiment_2_100_10_3_mod}}
    \subfloat[]{\input{images/experiment_2_200_20_3_mod}}
    \caption{Experiments for Setting \#2.2: (a) $\Tmax = 100$, $\alpha = 10$, (b) $\Tmax = 200$, $\alpha = 20$.} \label{fig:late}
\end{figure}

The corresponding results are provided in Figure~\ref{fig:late}.
They are in line with the ones of Setting \#2.

\paragraph{Setting \#2.3}
Finally, in this experiment, the setting is the same as the one in Setting \#2, except that the reward distributions are randomly chosen.

Specifically, the distribution sampled used in the experiments are:
\begin{itemize}
    \item $\Tmax = 100$, $\alpha = 10$:
    \begin{align*}
        &\bm{a}^i=[7,7,1,5,9,8,7,5,8,6]; \\ &\bm{b}^i=[10,4,9,3,5,3,2,10,5,9];
    \end{align*}
    \item $\Tmax = 200$, $\alpha = 20$:
\begin{align*}
    &\bm{a}^i=[10,3,5,2,2,6,8,9,2,6,7,6,10,4,9,8,8,9,5,1]; \\ &\bm{b}^i=[9,1,2,7,1,10,8,6,4,6,2,4,10,4,4,3,9,8,2,2];
\end{align*}
    \item $\Tmax = 100$, $\alpha = 50$:
\begin{align*}
    \bm{a}^i=[&6,9,8,2,5,9,5,2,9,6,9,4,10,9,10,5,8,2,10,7,6,10,4,5,3,4,3,1,10,5,8,2,2,3,3,\\
    &1,2,9,7,9,5,9,4,4,10,7,10,5,8,8]; \\ 
    \bm{b}^i=[&6,2,6,10,2,8,10,6,4,4,1,5,2,4,6,3,6,7,1,2,3,4,1,10,9,10,2,1,2,4,10,10,2,7,2,\\
    &6,2,1,10,1,4,3,2,8,4,1,1,9,7,10];
\end{align*}
    \item $\Tmax = 200$, $\alpha = 100$:
\begin{align*}
    \bm{a}^i=[&2,5,2,4,2,5,6,7,3,1,9,8,1,10,2,7,4,5,6,8,10,3,4,1,3,3,6,9,5,2,10,8,3,1,8,7,\\
    &10,9,5,6,7,5,3,9,1,8,2,6,1,9,5,3,4,8,6,10,5,6,10,10,3,5,7,7,2,1,10,4,6,3,4,\\
    &4,8,7,10,7,1,7,10,7,1,3,8,2,5,3,8,9,8,9,10,1,1,8,6,5,8,1,7,4]; \\ 
    \bm{b}^i=[&9,2,3,1,7,7,6,1,4,1,1,9,10,2,4,2,10,4,5,5,3,2,8,7,2,1,5,8,2,5,3,9,6,2,3,5,1,\\
    &1,1,4,5,9,6,6,10,1,10,8,8,7,6,9,3,4,7,10,5,1,3,3,5,6,6,6,2,6,10,1,1,5,3,3,\\
    &10,5,6,7,9,3,5,2,8,4,1,5,3,9,2,5,7,6,5,7,2,2,9,8,8,6,6,2].
\end{align*}
\end{itemize}

\begin{figure}[t!]
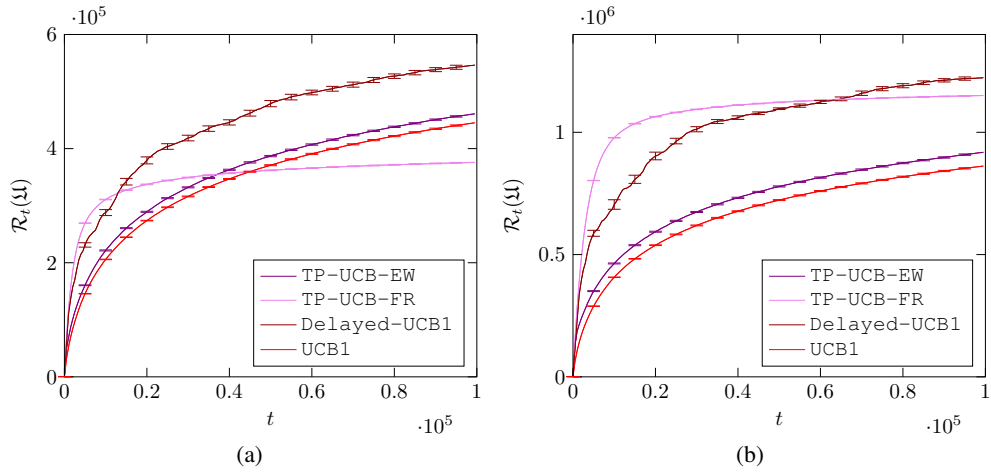

    \centering
    \subfloat[]{\input{images/experiment_2_100_10_4_mod}}
    \subfloat[]{\input{images/experiment_2_200_20_4_mod}}
    \caption{Experiments for Setting \#2.3: (a) $\Tmax = 100$, $\alpha = 10$, (b) $\Tmax = 200$, $\alpha = 20$.} \label{fig:rand}
\end{figure}

The corresponding results are provided in Figure~\ref{fig:rand}.
They are in line with the ones of Setting \#2.

\paragraph{Summary for Setting \#2}
The overall results for the previous setting \#2, \#2.1, \#2.2, and \#2.3 are reported in Table~\ref{t:summary_exp2_100_10}, \ref{t:summary_exp2_200_20}, \ref{t:summary_exp2_100_50}, \ref{t:summary_exp2_200_100}.

\begin{table}[t!]
\caption{Summary of result for setting \#2, $\Tmax = 100$, $\alpha = 10$.} \label{t:summary_exp2_100_10}
\begin{center}
	\begin{tabular}{|c|c|c|c|c|c|}
	    \hline
		$\Tmax$ & $\alpha$ & Scenario & Learner & Regret & Confidence Interval \\
		\hline
		 100 & 10  & 1 & \texttt{TP-UCB-FR} & 379407.7536 & 641.3890868 \\
		\hline
		 100 & 10  & 1 & \texttt{TP-UCB-EW} & 476211.7734 & 1379.593546\\
		\hline
		 100 & 10  & 1 & \texttt{Delayed-UCB1} & 550020.3093 & 3383.218936\\
		\hline
		 100 & 10  & 1 & \texttt{UCB1} & 461295.3133 & 1198.377002\\
		\hline
		\hline
	    100 & 10  & 2 & \texttt{TP-UCB-FR} &	378590.4996	& 1444.810301 \\
	    \hline
		100 & 10  & 2 &\texttt{TP-UCB-EW} &	478543.3454 &	3282.169025\\
        \hline
        100 & 10  & 2 & \texttt{Delayed-UCB1} & 556264.2577 & 4563.491842\\
        \hline
        100 & 10  & 2 & \texttt{UCB1} & 464045.2915	& 3127.506071\\
        \hline
        \hline
        100 & 10  & 3 & \texttt{TP-UCB-FR} &	377928.2537	& 550.2470147 \\
	    \hline
		100 & 10  & 3 &\texttt{TP-UCB-EW} &	477050.7314	& 1370.65113\\
        \hline
        100 & 10  & 3 & \texttt{Delayed-UCB1} & 552254.3013 &	2871.253395\\
        \hline
        100 & 10  & 3 & \texttt{UCB1} & 462051.9847 &	1022.873814\\
        \hline
        \hline
        100 & 10  & 4 & \texttt{TP-UCB-FR} & 376004.9497 &	713.1333679 \\
		\hline
		 100 & 10  & 4 & \texttt{TP-UCB-EW} & 461523.0728 &	1159.826331\\
		\hline
		 100 & 10  & 4 & \texttt{Delayed-UCB1} & 546401.0207	& 3116.186928\\
		\hline
		 100 & 10  & 4 & \texttt{UCB1} & 445761.5334	& 1160.681727\\
		\hline
    \end{tabular}
\end{center}
\end{table}

\begin{table}[t!]
\caption{Summary of result for setting \#2, $\Tmax = 200$, $\alpha = 20$.}
\label{t:summary_exp2_200_20}
\begin{center}
	\begin{tabular}{|c|c|c|c|c|c|}
	    \hline
		$\Tmax$ & $\alpha$ & Scenario & Learner & Regret & Confidence Interval \\
		\hline
		 200 & 20  & 1 & \texttt{TP-UCB-FR} & 1161392.507 & 653.9898656\\
		\hline
		 200 & 20  & 1 & \texttt{TP-UCB-EW} & 969119.3579 & 2376.133933\\
		\hline
		 200 & 20  & 1 & \texttt{Delayed-UCB1} & 1215396.1 & 11238.84718\\
		\hline
		 200 & 20  & 1 & \texttt{UCB1} & 921857.7185	& 1262.074342\\
		\hline
		\hline
	    200 & 20  & 2 & \texttt{TP-UCB-FR} & 1159038.888 & 	1855.393219 \\
	    \hline
		200 & 20  & 2 &\texttt{TP-UCB-EW} &976387.8607	&4103.793005	\\
        \hline
        200 & 20  & 2 & \texttt{Delayed-UCB1} & 1214717.526	& 12958.26024\\
        \hline
        200 & 20  & 2 & \texttt{UCB1} &922123.0453 &3911.196296	\\
        \hline
        \hline
        200 & 20  & 3 & \texttt{TP-UCB-FR} & 1158406.886 & 719.1511692\\
	    \hline
		200 & 20  & 3 &\texttt{TP-UCB-EW} & 971023.1429 &2128.831649	\\
        \hline
        200 & 20  & 3 & \texttt{Delayed-UCB1} & 1225998.654	& 12586.53841\\
        \hline
        200 & 20  & 3 & \texttt{UCB1} & 922097.5566	& 1084.342302\\
        \hline
        \hline
        200 & 20  & 4 & \texttt{TP-UCB-FR} & 1150596.776 & 1373.38433 \\
		\hline
		200 & 20  & 4 & \texttt{TP-UCB-EW} & 919231.1795 & 2971.38115\\
		\hline
		200 & 20  & 4 & \texttt{Delayed-UCB1} & 1224143.761 & 6816.6797\\
		\hline
		200 & 20  & 4 & \texttt{UCB1} & 863043.4276 & 2568.233259 \\
		\hline
    \end{tabular}
\end{center}
\end{table}

\begin{table}[t!]
\caption{Summary of result for setting \#2, $\Tmax = 100$, $\alpha = 50$.}
\label{t:summary_exp2_100_50}
\begin{center}
	\begin{tabular}{|c|c|c|c|c|c|}
	    \hline
		$\Tmax$ & $\alpha$ & Scenario & Learner & Regret & Confidence Interval \\
		\hline
		 100 & 50  & 1 & \texttt{TP-UCB-FR} &  280850.7628 & 200.0363298\\
		\hline
		 100 & 50  & 1 & \texttt{TP-UCB-EW} & 470206.8356 &	610.8394845\\
		\hline
		 100 & 50  & 1 & \texttt{Delayed-UCB1} & 555004.3727 & 3611.482174\\
		\hline
		 100 & 50  & 1 & \texttt{UCB1} & 461125.7678 & 433.1909748\\
		\hline
		\hline
	    100 & 50  & 2 & \texttt{TP-UCB-FR} & 280469.8885 & 600.1158378	 \\
	    \hline
		100 & 50  & 2 &\texttt{TP-UCB-EW} &	470948.6985	& 1810.491059\\
        \hline
        100 & 50  & 2 & \texttt{Delayed-UCB1} & 551713.5918 & 3167.855141\\
        \hline
        100 & 50  & 2 & \texttt{UCB1} & 460454.4842&	1535.465475\\
        \hline
        \hline
        100 & 50  & 3 & \texttt{TP-UCB-FR} & 	280432.6875	& 194.6246275\\
	    \hline
		100 & 50  & 3 &\texttt{TP-UCB-EW} &	470851.5341&	678.1378134\\
        \hline
        100 & 50  & 3 & \texttt{Delayed-UCB1} & 552354.8852&	2784.797814\\
        \hline
        100 & 50  & 3 & \texttt{UCB1} & 461262.8902&	406.9041603\\
        \hline
        \hline
        100 & 50  & 4 & \texttt{TP-UCB-FR} & 277350.6683  & 357.2049513\\
		\hline
		 100 & 50  & 4 & \texttt{TP-UCB-EW} & 431428.2109 &	845.9105653\\
		\hline
		 100 & 50  & 4 & \texttt{Delayed-UCB1} & 533550.167 &	6134.964191\\
		\hline
		 100 & 50  & 4 & \texttt{UCB1} & 419308.3464	& 840.25097\\
		\hline
    \end{tabular}
\end{center}
\end{table}

\begin{table}[t!]
\caption{Summary of result for setting \#2, $\Tmax = 200$, $\alpha = 100$.}
\label{t:summary_exp2_200_100}
\begin{center}
	\begin{tabular}{|c|c|c|c|c|c|}
	    \hline
		$\Tmax$ & $\alpha$ & Scenario & Learner & Regret & Confidence Interval \\
		\hline
		 200 & 100  & 1 & \texttt{TP-UCB-FR} & 998723.9102 & 348.3923308\\
		\hline
		 200 & 100  & 1 & \texttt{TP-UCB-EW} & 962166.9976 & 1574.53646\\
		\hline
		 200 & 100  & 1 & \texttt{Delayed-UCB1} & 1217054.205 & 13791.12121\\
		\hline
		 200 & 100  & 1 & \texttt{UCB1} & 922801.461	& 681.1463488\\
		\hline
		\hline
	    200 & 100  & 2 & \texttt{TP-UCB-FR} & 997866.0232 & 1163.306506	 \\
	    \hline
		200 & 100  & 2 &\texttt{TP-UCB-EW} &962888.2947	&2886.588981	\\
        \hline
        200 & 100  & 2 & \texttt{Delayed-UCB1} & 1223555.271	& 13076.51935\\
        \hline
        200 & 100  & 2 & \texttt{UCB1} & 924666.3352 &	1936.282782\\
        \hline
        \hline
        200 & 100  & 3 & \texttt{TP-UCB-FR} & 995734.719 & 386.1528975\\
	    \hline
		200 & 100  & 3 &\texttt{TP-UCB-EW} & 962419.0355 &	1671.591765\\
        \hline
        200 & 100  & 3 & \texttt{Delayed-UCB1} & 1224181.588	& 14560.25523\\
        \hline
        200 & 100  & 3 & \texttt{UCB1} & 923018.9128	& 593.7216922\\
        \hline
        \hline
        200 & 100  & 4 & \texttt{TP-UCB-FR} & 996058.5901 &	681.2301995 \\
		\hline
		200 & 100  & 4 & \texttt{TP-UCB-EW} & 937032.8774 &	1815.90584\\
		\hline
		200 & 100  & 4 & \texttt{Delayed-UCB1} & 1214671.825 &	12459.63383\\
		\hline
		200 & 100  & 4 & \texttt{UCB1} & 893569.8466 &	1098.403796\\
		\hline
    \end{tabular}
\end{center}
\end{table}

\clearpage

\paragraph{Setting \#4}

\begin{figure}[t!]
    \centering
    \input{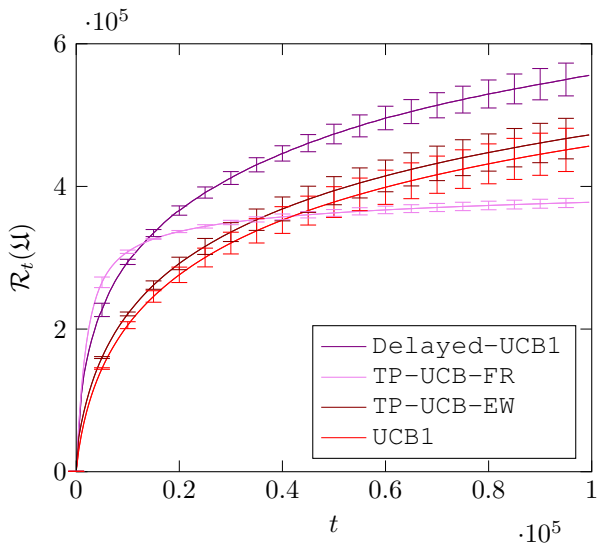}
    \caption{Experiments for Setting \# 4: $\Tmax = 100$, $\alpha = 10$} \label{fig:exp4}
\end{figure}

In this setting, each arm is described by a maximum reward $\overline{R}^i = \Tmax \cdot i$, and two vectors $\bm{a}^i=\mleft[a^i_{1},\ldots,a^i_{\alpha}\mright]$ and $\bm{b}^i=\mleft[b^i_{1},\ldots,b^i_{\alpha}\mright]$ of length $\alpha$.
The aggregated rewards $Z^i_{t,k}$ are distributed as $\mathcal{D}_k^i = \frac{\overline{R}^i}{\alpha}Beta(a_k^i,b_k^i)$, $\forall k \in [\alpha]$.
In this experiment, we fix $\Tmax = 100$, $\alpha = 10$, $T = 10^5$, and we design ten scenarios differing in the vectors $\bm{a}^i$ and $\bm{b}^i$.
The parameters characterizing such randomly generated scenarios are reported in Table~\ref{t:env_4}.
The results for each scenario are averaged over $50$ independent runs.
In Figure~\ref{fig:exp4}, we provide the average result over the $10$ scenarios, with whiskers corresponding to $95\%$ confidence intervals.

\begin{table}[t!]
\caption{Parameters used in Setting \#4.}
\label{t:env_4}
\begin{center}
	\begin{tabular}{|c|c|c|}
	    \hline
		& $\bm{a}^i$ & $\bm{b}^i$ \\
		\hline
		Scenario 1 & [8, 2, 8, 7, 1, 5, 6, 3, 3, 10] & [7, 2, 2, 2, 4, 4, 1, 7, 1, 2]\\
		\hline
		Scenario 2 & [7, 9, 9, 5, 8, 8, 10, 4, 7, 2] & [6, 4, 5, 10, 3, 7, 4, 6, 2, 2]\\
		\hline
		Scenario 3 & [1, 9, 8, 4, 2, 8, 7, 5, 4, 1] & [4, 10, 3, 2, 4, 8, 7, 6, 9, 3]\\
		\hline
		Scenario 4 & [2, 10, 8, 3, 10, 7, 7, 9, 8, 6] & [8, 8, 4, 9, 10, 4, 1, 6, 6, 6]\\
		\hline
		Scenario 5 & [1, 9, 3, 5, 10, 3, 7, 10, 5, 8] & [2, 2, 9, 1, 2, 4, 3, 1, 5, 1]\\
		\hline
		Scenario 6 & [8, 6, 3, 3, 8, 6, 9, 7, 9, 9] & [1, 10, 2, 9, 10, 2, 7, 4, 5, 9]\\
		\hline
		Scenario 7 & [10, 7, 8, 7, 10, 10, 4, 1, 1, 3] & [5, 9, 10, 5, 6, 2, 8, 5, 5, 7]\\
		\hline
		Scenario 8 & [7, 7, 1, 3, 3, 4, 5, 6, 1, 1] & [8, 7, 3, 8, 10, 2, 3, 6, 7, 1]\\
		\hline
		Scenario 9 & [10, 8, 7, 8, 1, 2, 8, 3, 1, 1] & [10, 10, 3, 6, 2, 9, 6, 4, 7, 8]\\
		\hline
		Scenario 10 & [2, 1, 10, 8, 10, 6, 2, 10, 5, 3] & [7, 5, 2, 9, 4, 1, 7, 8, 6, 4]\\
		\hline
	\end{tabular}
\end{center}
\end{table}

Figure~\ref{fig:exp4} shows an aggregated result on the pseudo-regret $\mathcal{R}_t(\mathfrak{U})$ for the analysed algorithms.
Even over randomly generated scenarios we see that the proposed method are able to provide a significant improvement over the \texttt{Delayed-UCB1} algorithm.
Moreover, consistently the \texttt{TP-UCB-FR} algorithm result to be the best one at the end of the analysed time horizon $T = 10^5$.
Conversely, for shorter time horizon ($T \leq 0.35 \cdot 10^5$) the algorithm performing the best among the ones for the TP-MAB setting is the \texttt{TP-UCB-EW}, which strengthen the idea that this algorithm is better suited for shorter time horizons.

\begin{figure}[t!]
	\centering
	\includegraphics[width=0.9\textwidth]{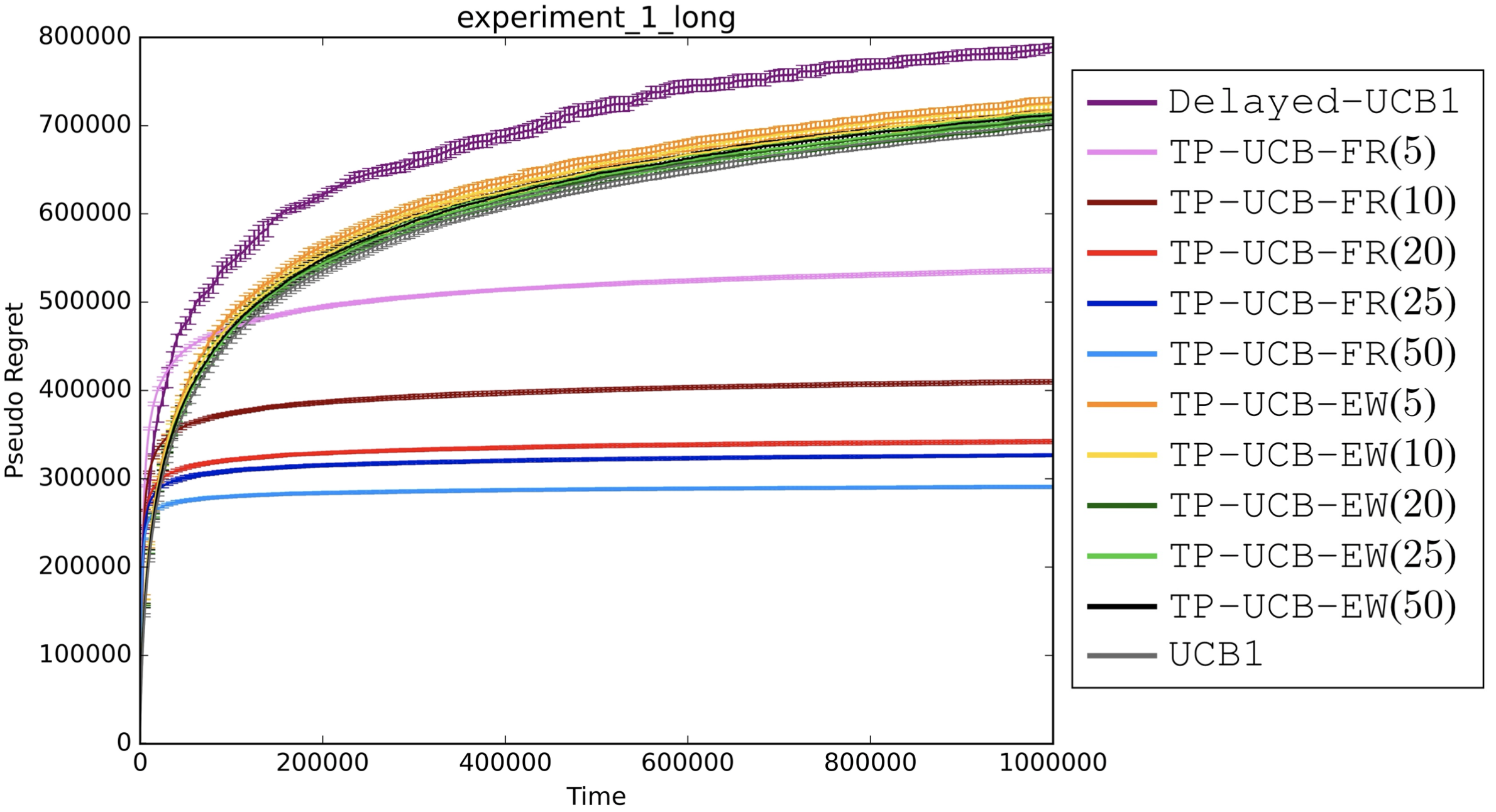}
	\caption{Experiments for Setting \# 5: $\Tmax = 100$, $\alpha = 20$} \label{fig:explong}
\end{figure}

\paragraph{Setting \#5}
Finally, we provide an experiment over a longer time horizon of $T = 10^6$ in the same configuration depicted by Setting \#1.
The pseudo-regret over time for this experiment is provided in Figure~\ref{fig:explong}.
Let us focus on the regret of \texttt{TP-UCB-FR}(20), \emph{i.e.}, the \texttt{TP-UCB-FR} algorithm where parameter $\alpha$ corresponds to the one of the environment, and compare it with the regret of \texttt{Delayed-UCB1}.
The regret of \texttt{TP-UCB-FR}(20) (red line) has a slower growth w.r.t.~\texttt{Delayed-UCB1} (purple line), and, consequently, the difference in terms of regret increases (logarithmically) over time.
The parameter influencing the regret of \texttt{TP-UCB-FR} is $\alpha$, which characterizes the specific setting we are tackling.
More specifically, if we fix the other parameters (e.g., $\tau_{\max}$) and increase the value of alpha, we have a proportional improvement in the upper bound of the regret of \texttt{TP-UCB-FR}.
Therefore, we expect to have an even larger improvement of our algorithm when the value of $\alpha$ is large.

\clearpage
\section{Real-world Applications of the TP-MAB Framework} \label{ap:examples}

In this section, we report some additional real-world examples which can be modeled through the TP-MAB setting. The following scenarios are characterized by the $\alpha$-smoothness property with different values of the $\alpha$ parameters.

\begin{example}[E-commerce]
An agent periodically receives a batch of identical items to sell on an e-commerce platform.
Every time a slot of $N$ items arrives, the agent decides a price $p_i$ to post on a website, which corresponds to the arm $i_t$ chosen for the round $t$.
The selected time horizon to sell the items, which are perishable, is one month.
Each day, the seller checks how many items have been ordered and collects the payments (\emph{i.e.}, rewards).
In this example, the maximum delay is $\Tmax = 30$ days, and one round is equal to $1$ day.
The upper bound on the cumulative reward is $\overline{R}^i = p_i N$.
Notice that the partial reward of each round is also upper bounded by $p_i N$.
This implies that the reward has no structure, and consequently the $\alpha$-smoothness in this setting holds with $\alpha = 1$.
\end{example}


\begin{example}[Lottery Ticket]
There are $K$ different lotteries to choose from.
Lottery $i \in [K]$ has $N$ winning scratch cards, each with a prize of $M$.
The probability to extract a winning ticket in lottery $i$ is $p_i$.
The player has to choose a lottery at each time step.
At each round, the player buys $n$ tickets and sequentially scratches them and observes the reward.
If $N = 1$ the total amount the player can win is $M$ and the reward is $1$-smooth.
Indeed, suppose that the first $n-1$ tickets are not winning.
This does not precludes the possibility of still gaining the maximum cumulative reward with the last ticket.
Conversely, if $N = n$ the total amount the player can win is $\overline{R}^i = N M$, and the reward is $n$-smooth.
More specifically, by scratching the first ticket, the player can get useful information on the cumulative reward if the reward is either zero or $M$.
If the player observed a zero reward so far, the maximum achievable cumulative reward becomes $(N-1) M$.
Conversely if the player observed a positive reward, the overall reward is in the interval $[M, N M]$.
\end{example}

\end{document}